\documentclass[journal]{IEEEtranTIE}
\usepackage{graphicx}
\usepackage{cite}
\usepackage{picinpar}
\usepackage{amsmath}
\usepackage{url}
\usepackage{flushend}
\usepackage[latin1]{inputenc}
\usepackage{colortbl}
\usepackage{soul}
\usepackage{multirow}
\usepackage{pifont}
\usepackage{color}
\usepackage{alltt}
\usepackage[hidelinks]{hyperref}
\usepackage{enumerate}
\usepackage{siunitx}
\usepackage{epstopdf}
\usepackage{pbox}
\usepackage{caption}
\usepackage[caption=false, font=footnotesize]{subfig}
\usepackage{amsfonts} 
\usepackage{amssymb}  
\usepackage{bm}       
\usepackage{booktabs}
\usepackage{algorithm}
\usepackage{algpseudocode}

\begin{document}
\title{L-SDPPO: Policy Optimization of Spiking Diffusion Policy for  Intra-vehicular Robotic Manipulation}

\author{
	\vskip 1em
	
	Liwen Zhang,
	Dong Zhou,
	Guanghui Sun,
	Yifei Zheng,
	Yuhui Hu,
	Kaihong Ouyang,
	\\ and Zuoquan Zhao
	
	\thanks{
		This work was supported in part by the Joint Funds of the National Natural Science Foundation of China through Grant U23A20346, and in part by the Fundamental and Interdisciplinary Disciplines Breakthrough Plan of the Ministry of Education of China through Grant No. JYB2025XDXM207.
		
		Liwen Zhang, Dong Zhou, Guanghui Sun, Yifei Zheng, Yuhui Hu, and Kaihong Ouyang are with the Department of Control Science and Engineering, Harbin Institute of Technology, Harbin 150001, China (e-mail: dongzhou@hit.edu.cn).
		
		Zuoquan Zhao is with the Department of Mechanical and Automation Engineering, The Chinese University of Hong Kong, Shatin, New Territories, Hong Kong SAR, China.
		
		Corresponding author: Dong Zhou.
	}
}

\maketitle
	
\begin{abstract}
Intra-vehicular robots in spacecraft help reduce astronaut workload and improve mission efficiency. Recent research focuses on using deep learning methods to achieve the acute control required for operations in these complex environments. However, objects exhibit unpredictable, unconstrained drift without gravitational damping. These factors demand robustness against complex multimodal action distributions. Diffusion policies (DP) can model these complex actions, but their iterative sampling process consumes too much energy for the limited power budgets of spacecraft. We therefore propose a low-energy intra-vehicular robotic manipulation framework, L-SDPPO, in which the Spiking Diffusion Policy (SDP) is optimized with a reinforcement learning (RL) algorithm. Furthermore, to address the insufficient perception of dynamic spatiotemporal features in microgravity, we propose the state-dependent latency injection (SDLI) mechanism, which mimics biological neural delays to dynamically regulate the timing of input information. Evaluation on five representative intra-vehicular daily tasks (e.g., hatch opening and precision container capping) shows that our method consistently achieves higher success rates and lower energy consumption, compared to the state-of-the-art robotic manipulation methods. These results demonstrate our method is a viable intra-vehicular robotic manipulation method. The open-source code and videos are
available at: \url{https://github.com/Dongzhou-1996/L-SDPPO.git}.
\end{abstract}


{}

\definecolor{limegreen}{rgb}{0.2, 0.8, 0.2}
\definecolor{forestgreen}{rgb}{0.13, 0.55, 0.13}
\definecolor{greenhtml}{rgb}{0.0, 0.5, 0.0}

\section{Introduction}
\IEEEPARstart{I}{ntra-vehicular robots} (IVRs) are essential for assisting astronauts and reducing workload during long-term space missions \cite{russell2006applying}. However, the microgravity environment renders free-floating objects highly susceptible to displacement, even under minimal interaction forces. Such sensitivity necessitates compliant grasping to prevent target drift, creating a demanding requirement for high-precision control \cite{li2022review}. Simultaneously, the stringent power limitations within space cabins require these sophisticated manipulation models to be exceptionally energy-efficient.

Early research on precision space operations relied heavily on model-based impedance control \cite{perez2018velocity}. These classical methods require extensive parameter tuning for specific hardware and often fail to generalize to unstructured tasks \cite{ma2023advances}. Data-driven approaches like reinforcement learning (RL) \cite{capra2024learning, hou2020data} and imitation learning (IL) \cite{li2025research_aca, wang2024novel} offer more flexibility, but they often suffer from low sample efficiency or struggle with multi-modal action distributions \cite{hua2021learning}.

\begin{figure}[t]
	\centering
	
	\subfloat[Task I: Hatch Opening]{
		\includegraphics[width=0.23\linewidth]{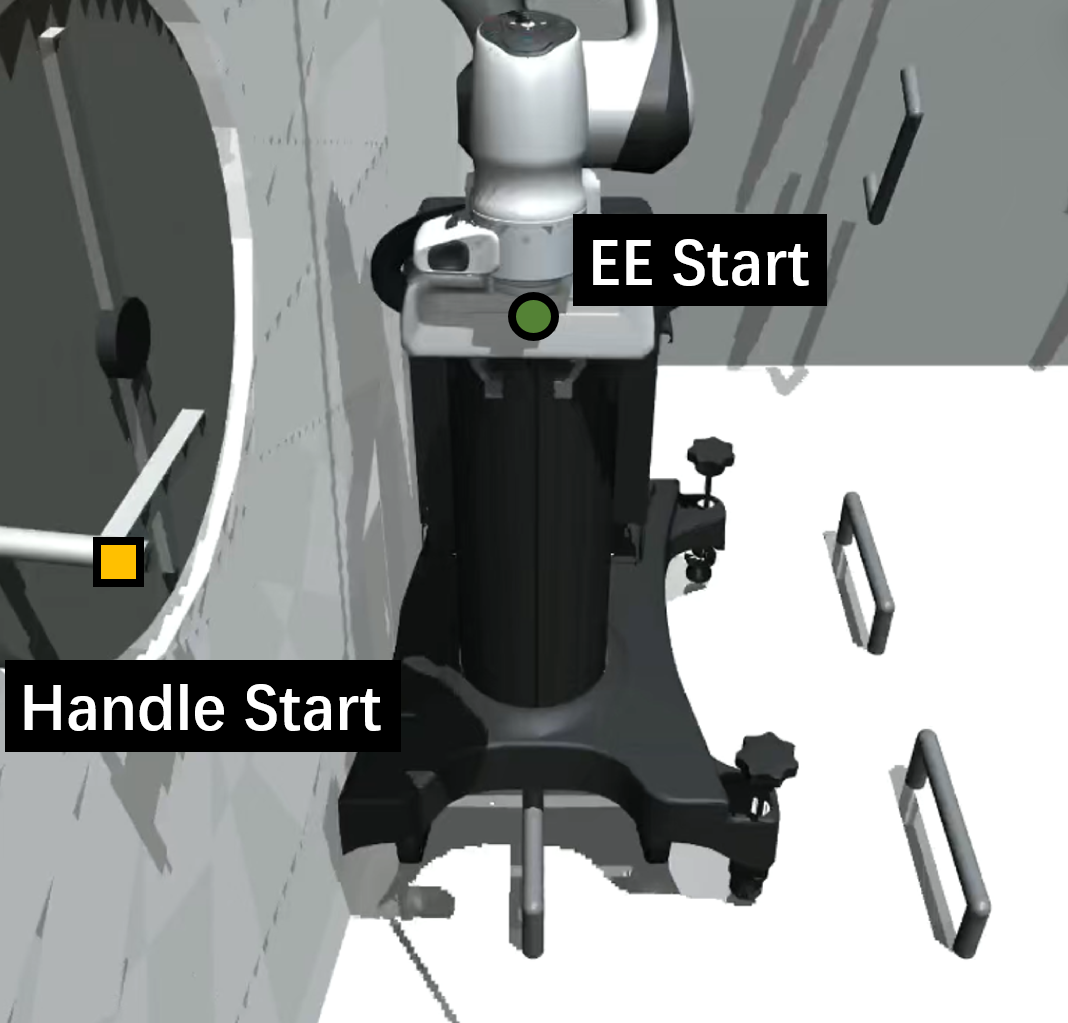} \hfill
		\includegraphics[width=0.23\linewidth]{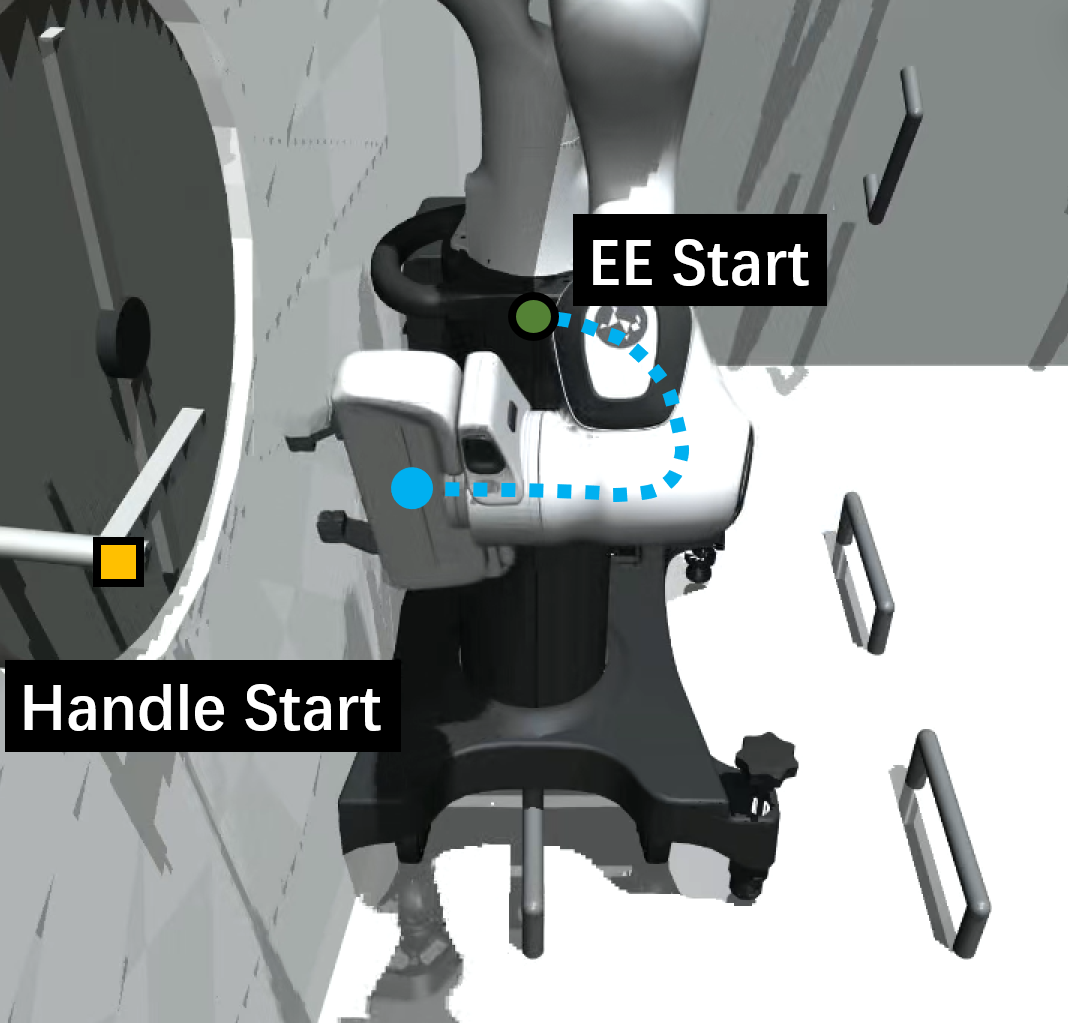} \hfill
		\includegraphics[width=0.23\linewidth]{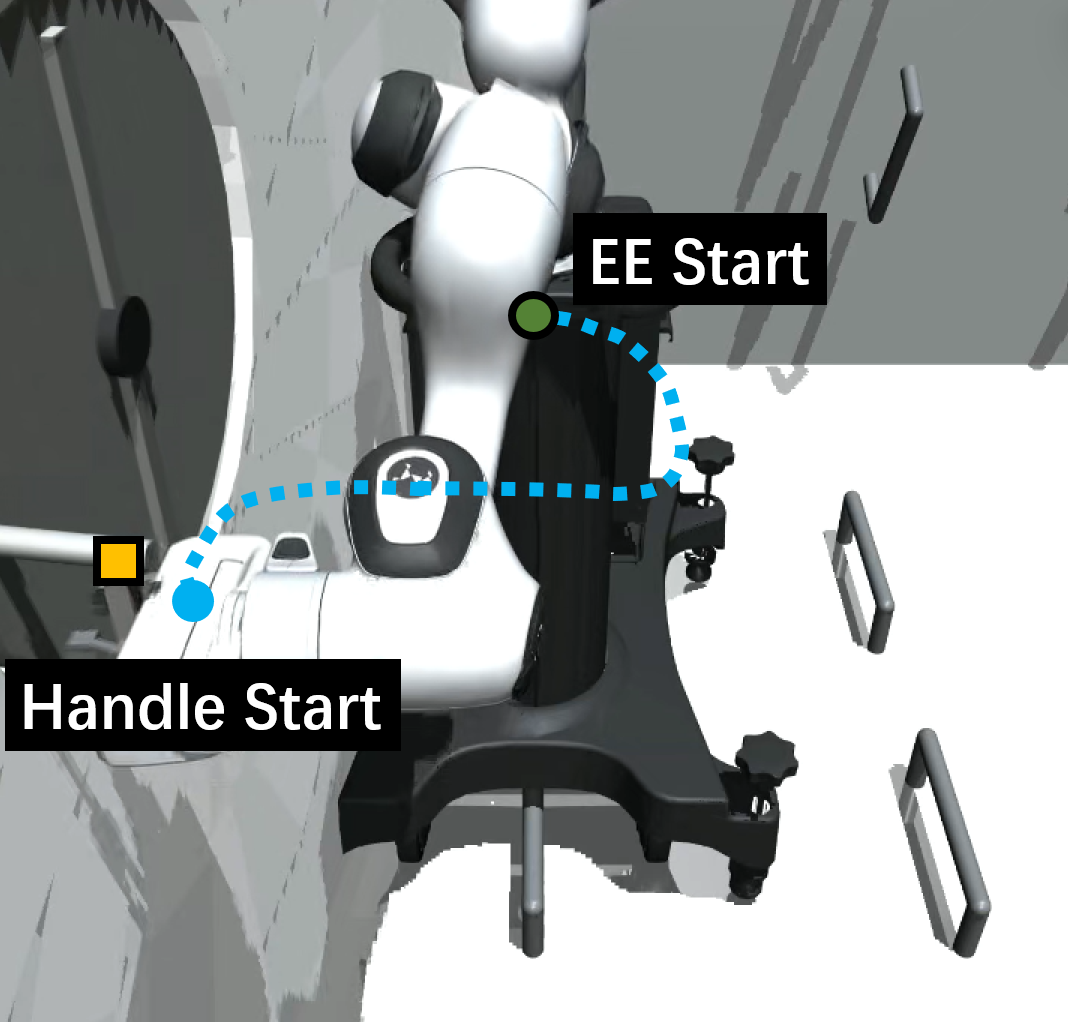} \hfill
		\includegraphics[width=0.23\linewidth]{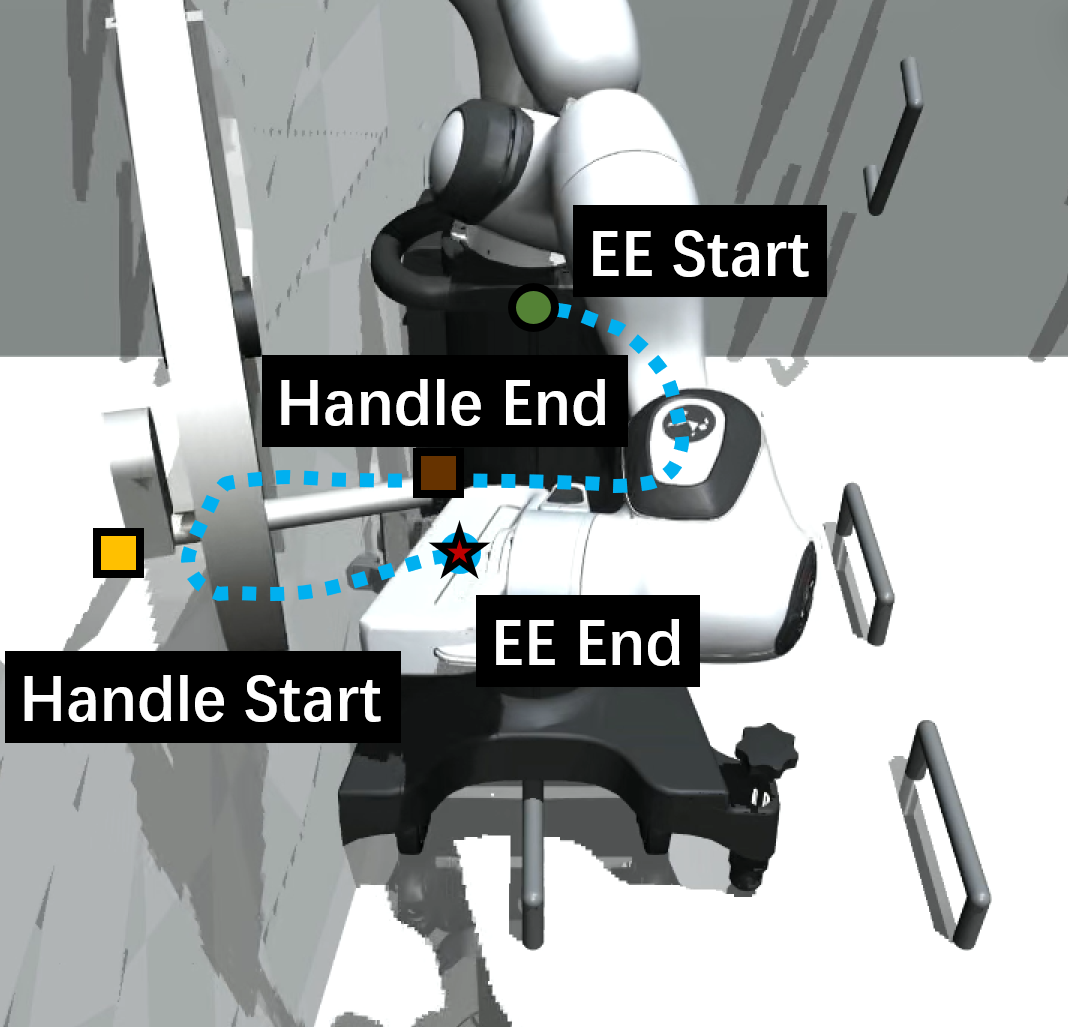}
	}

	\subfloat[Task II: Dynamic Tumbling Target Stowing in a Container]{
		\includegraphics[width=0.23\linewidth]{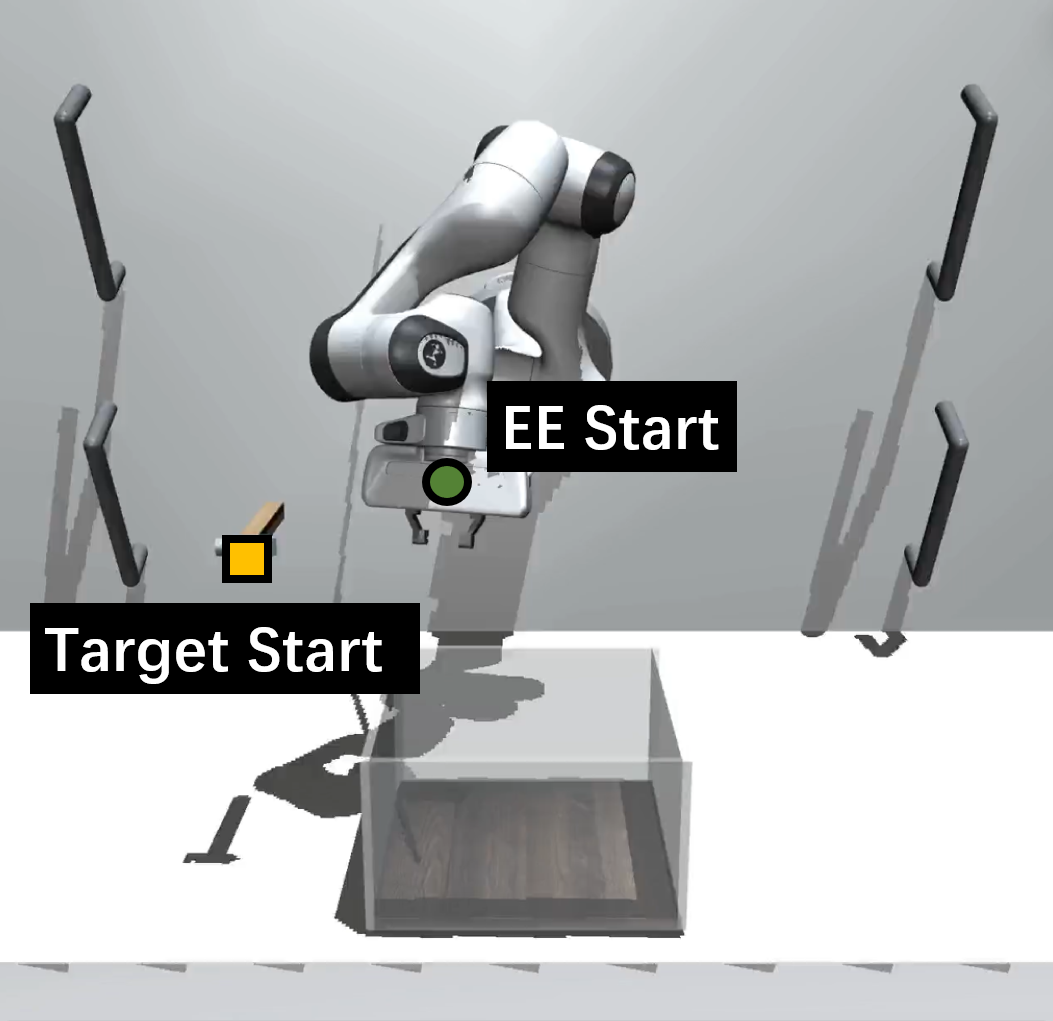} \hfill
		\includegraphics[width=0.23\linewidth]{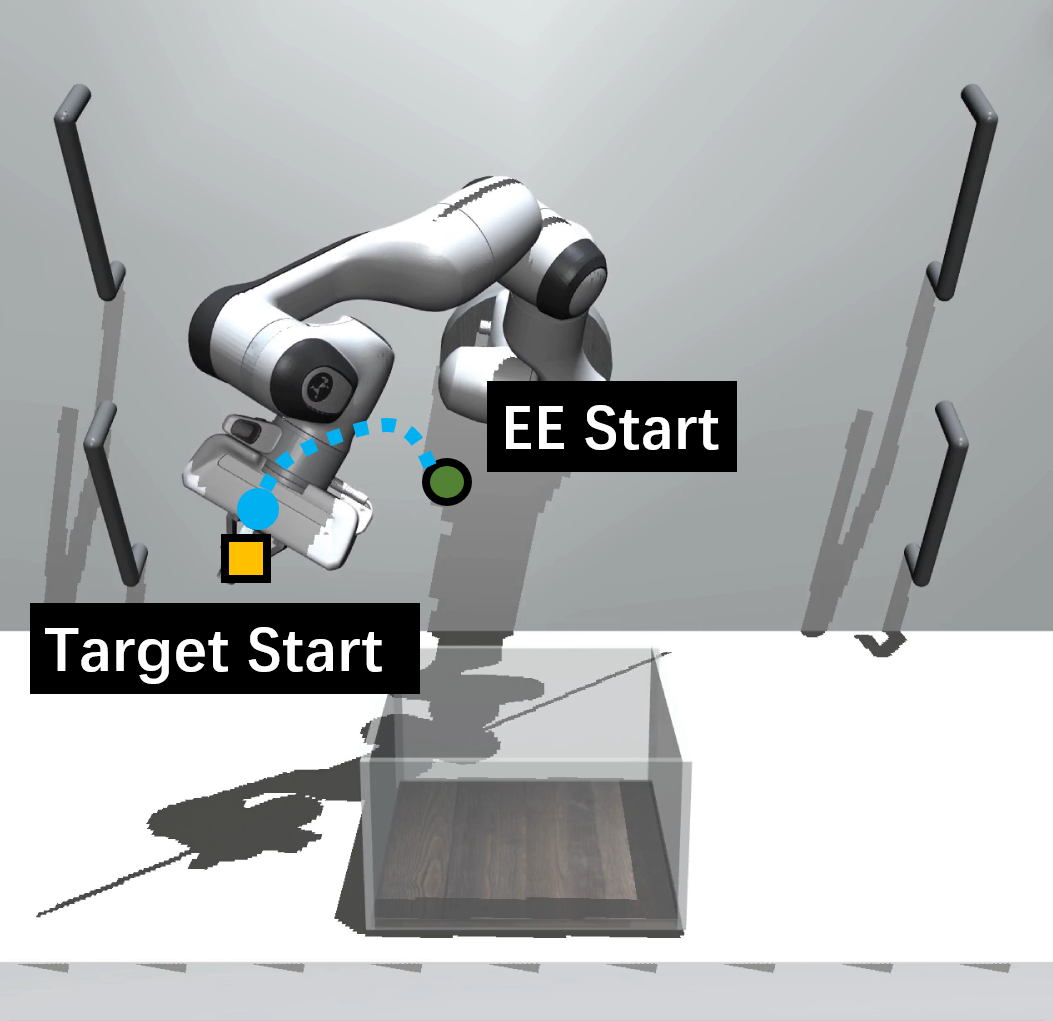} \hfill
		\includegraphics[width=0.23\linewidth]{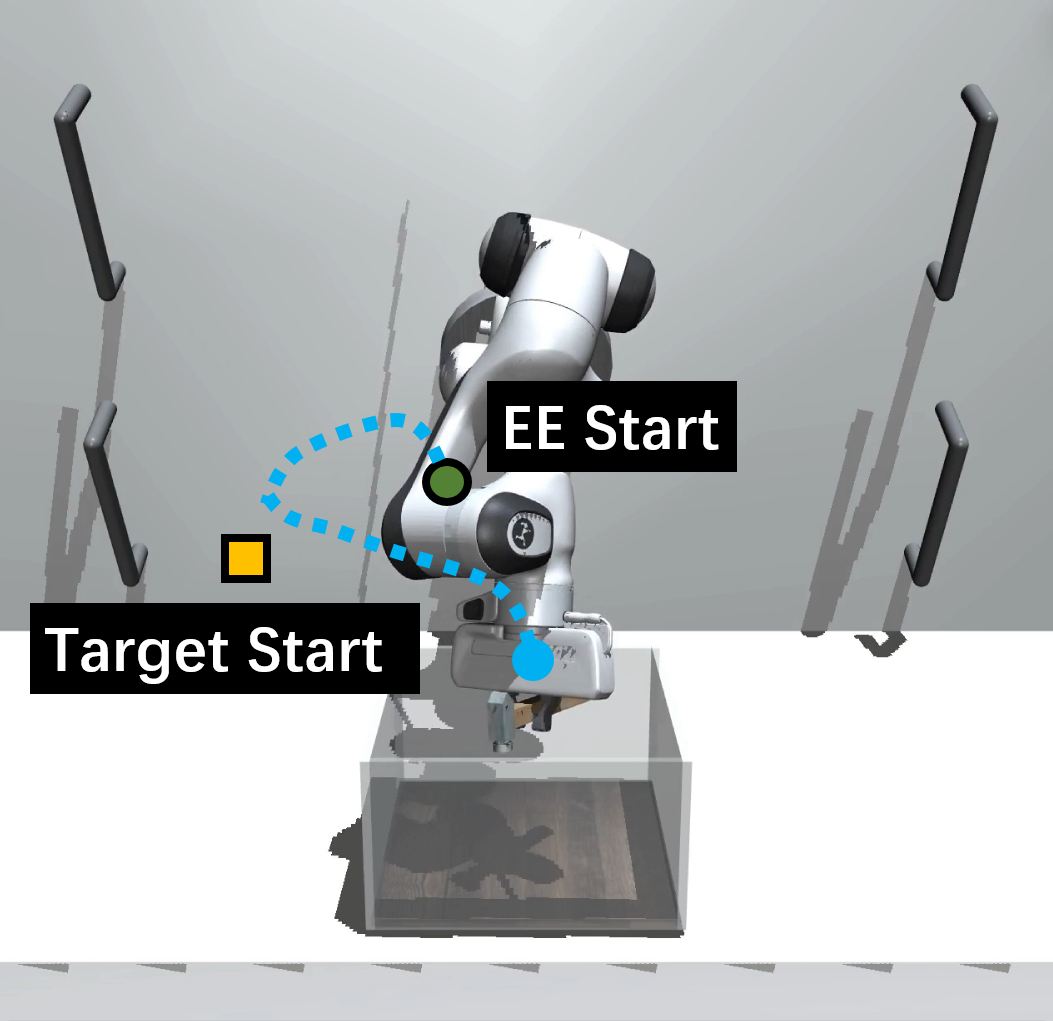} \hfill
		\includegraphics[width=0.23\linewidth]{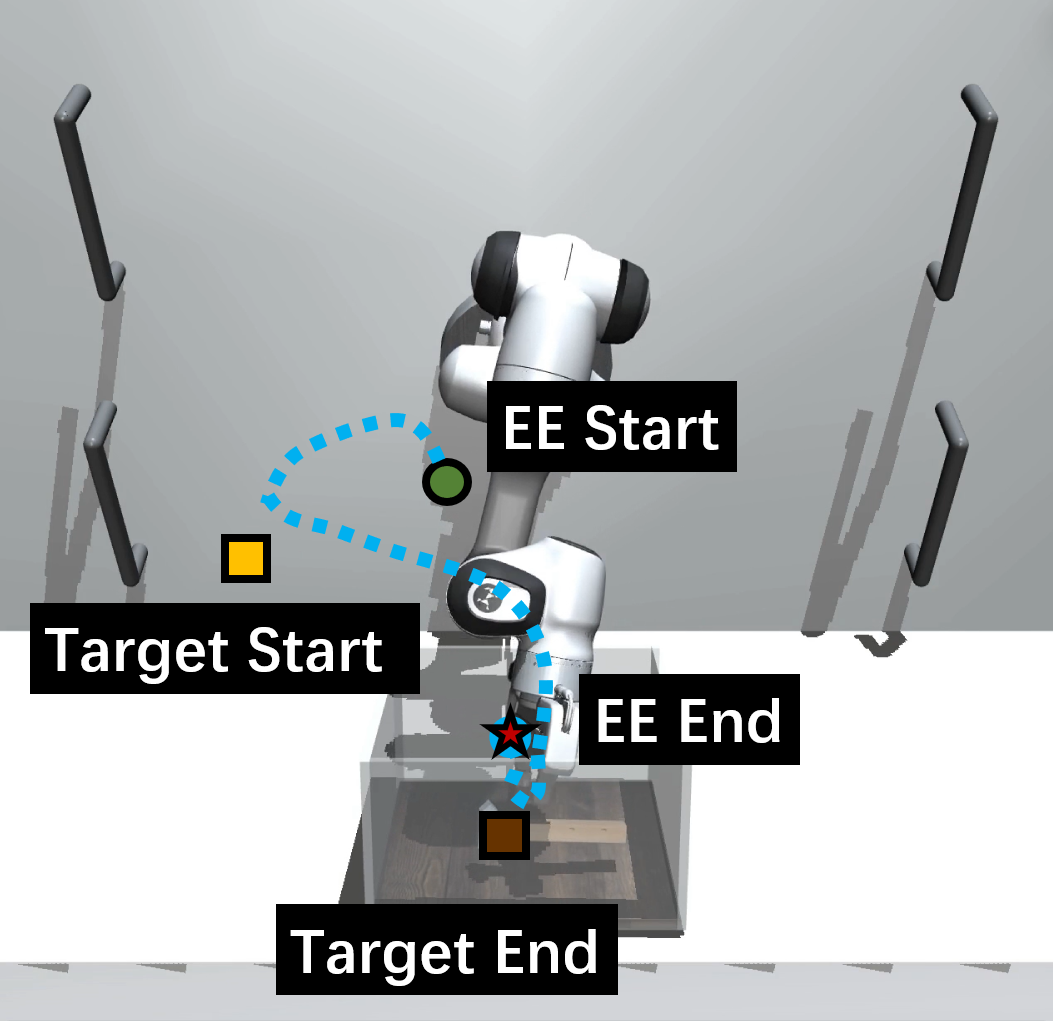}
	}

	\subfloat[Task III: Precision Container Capping]{
		\includegraphics[width=0.23\linewidth]{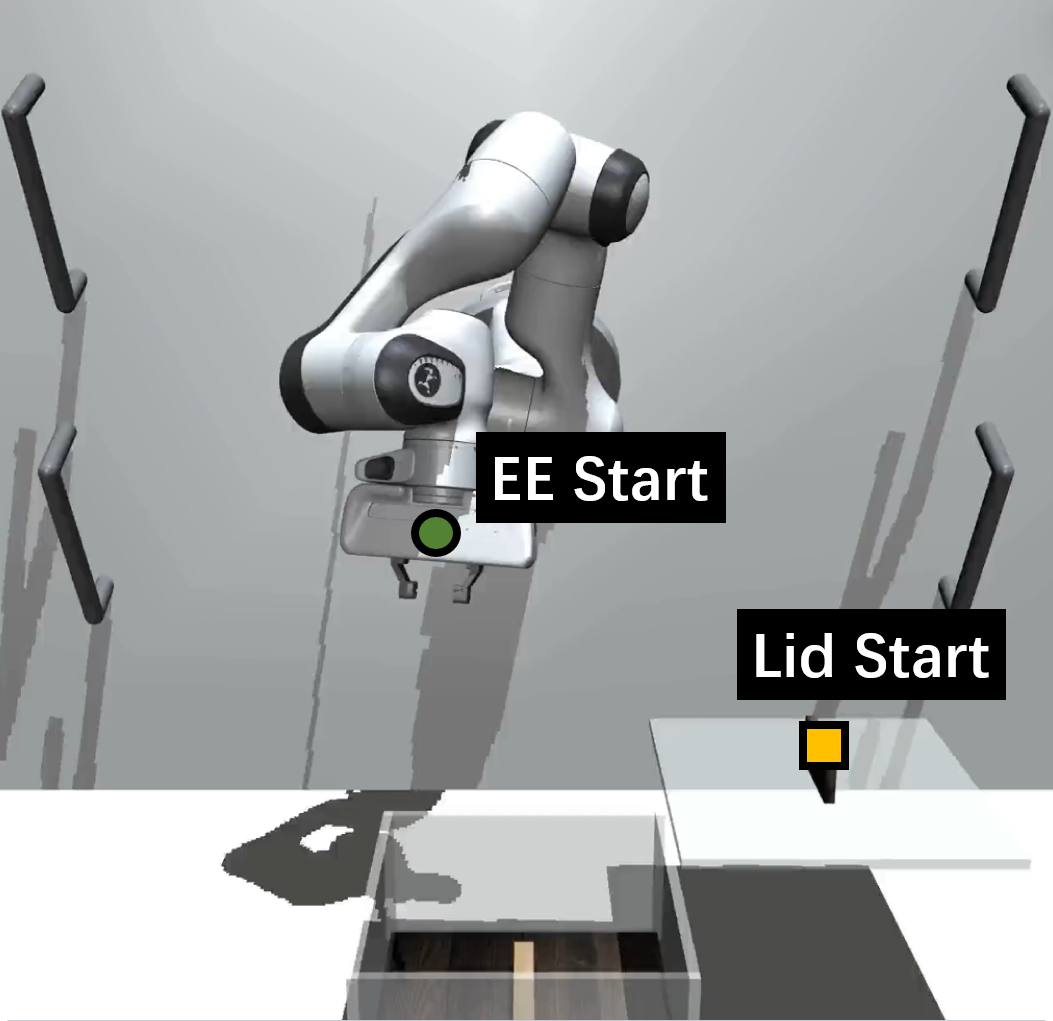} \hfill
		\includegraphics[width=0.23\linewidth]{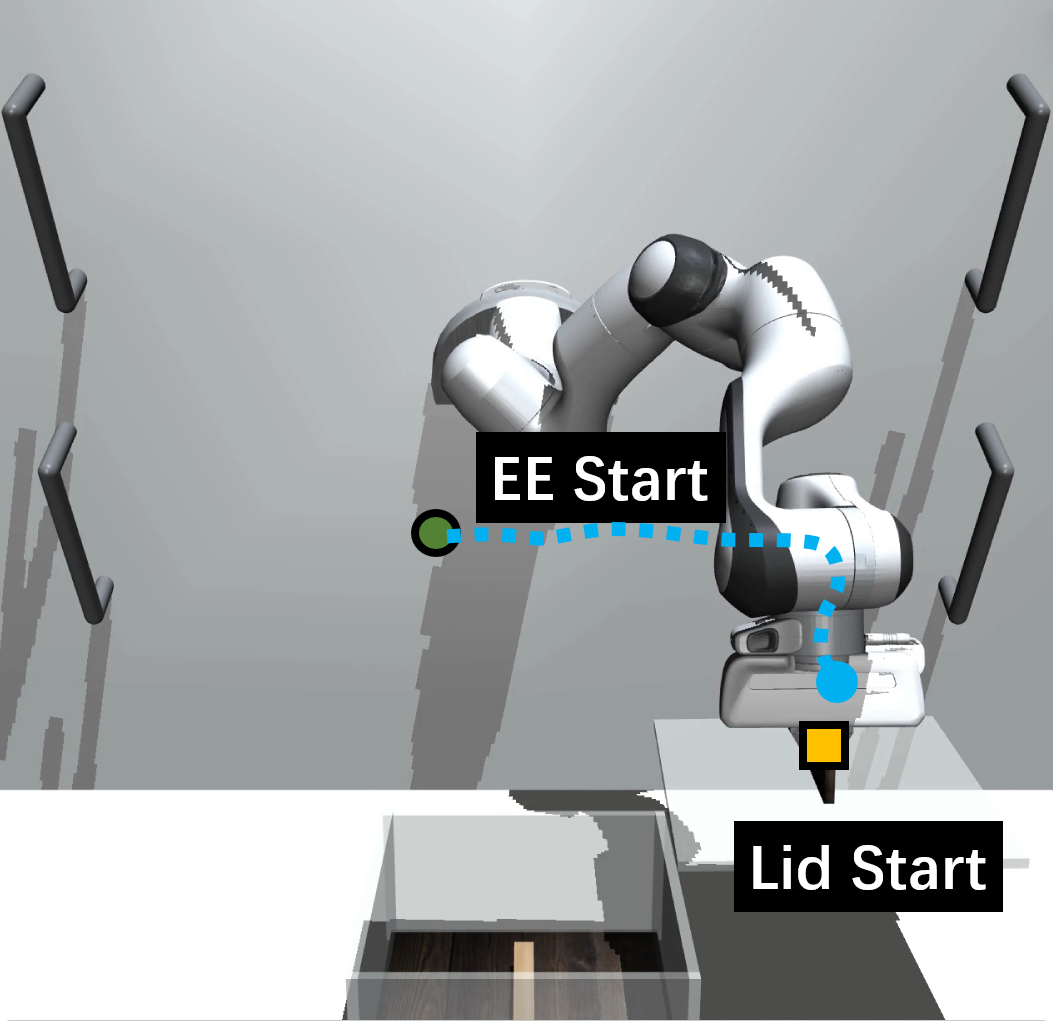} \hfill
		\includegraphics[width=0.23\linewidth]{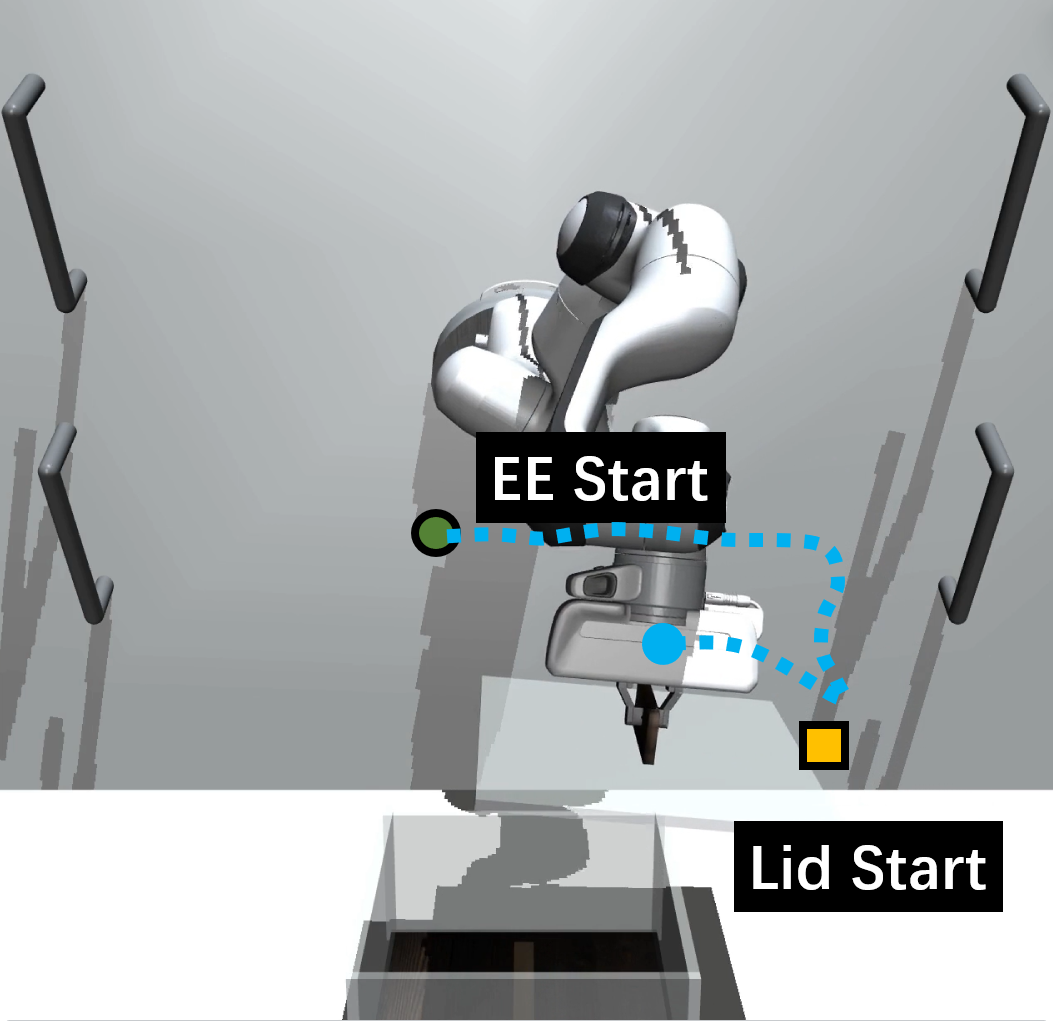} \hfill
		\includegraphics[width=0.23\linewidth]{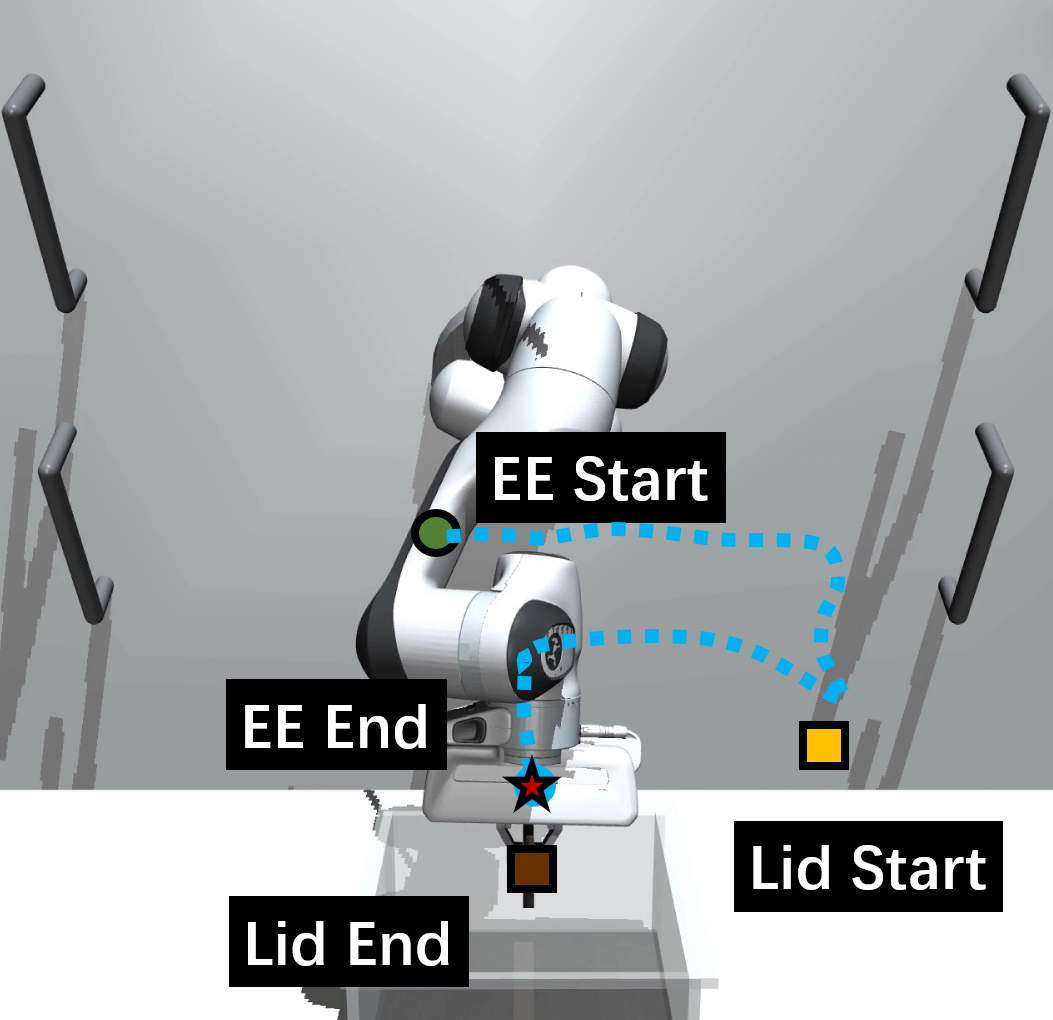}
	}

	\subfloat[Task IV: Sequential Panel Operation]{
		\includegraphics[width=0.23\linewidth]{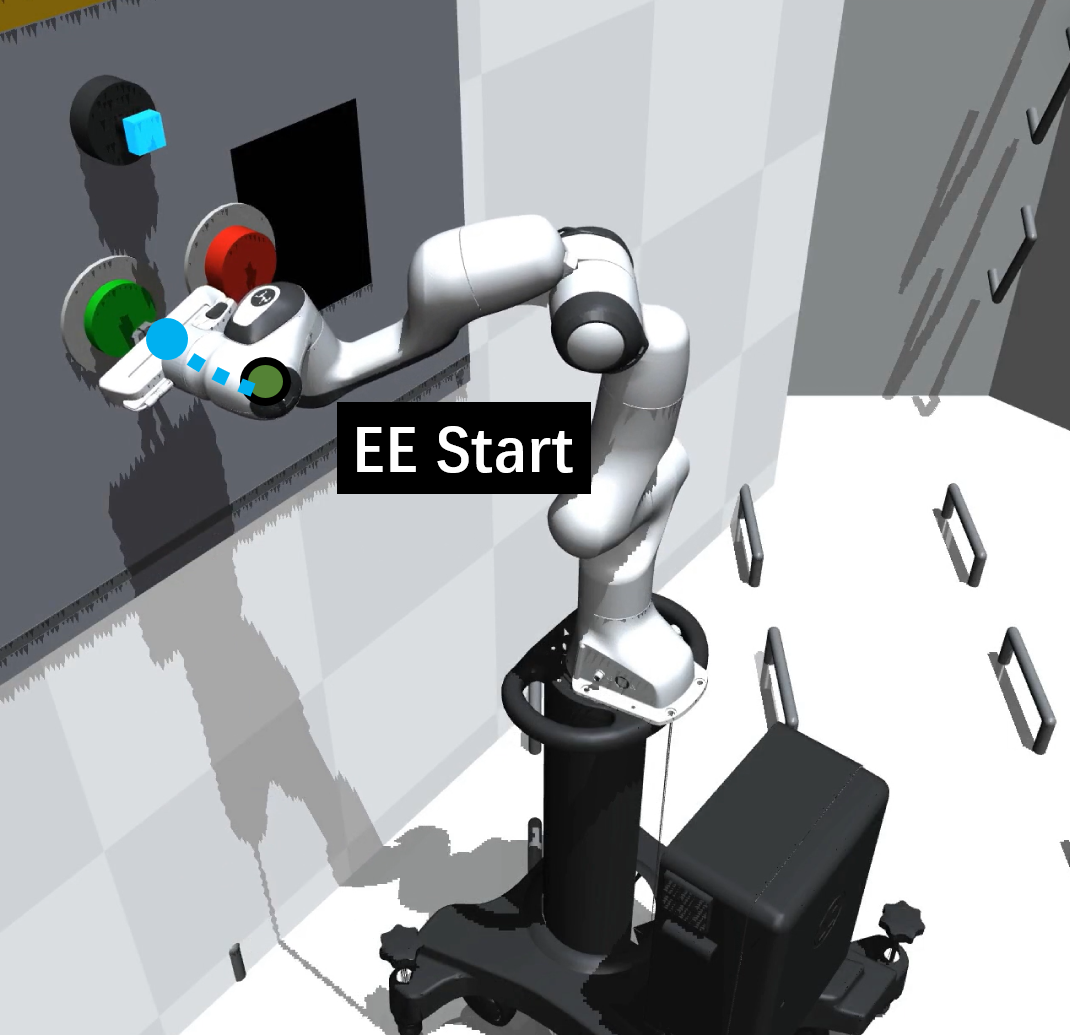} \hfill
		\includegraphics[width=0.23\linewidth]{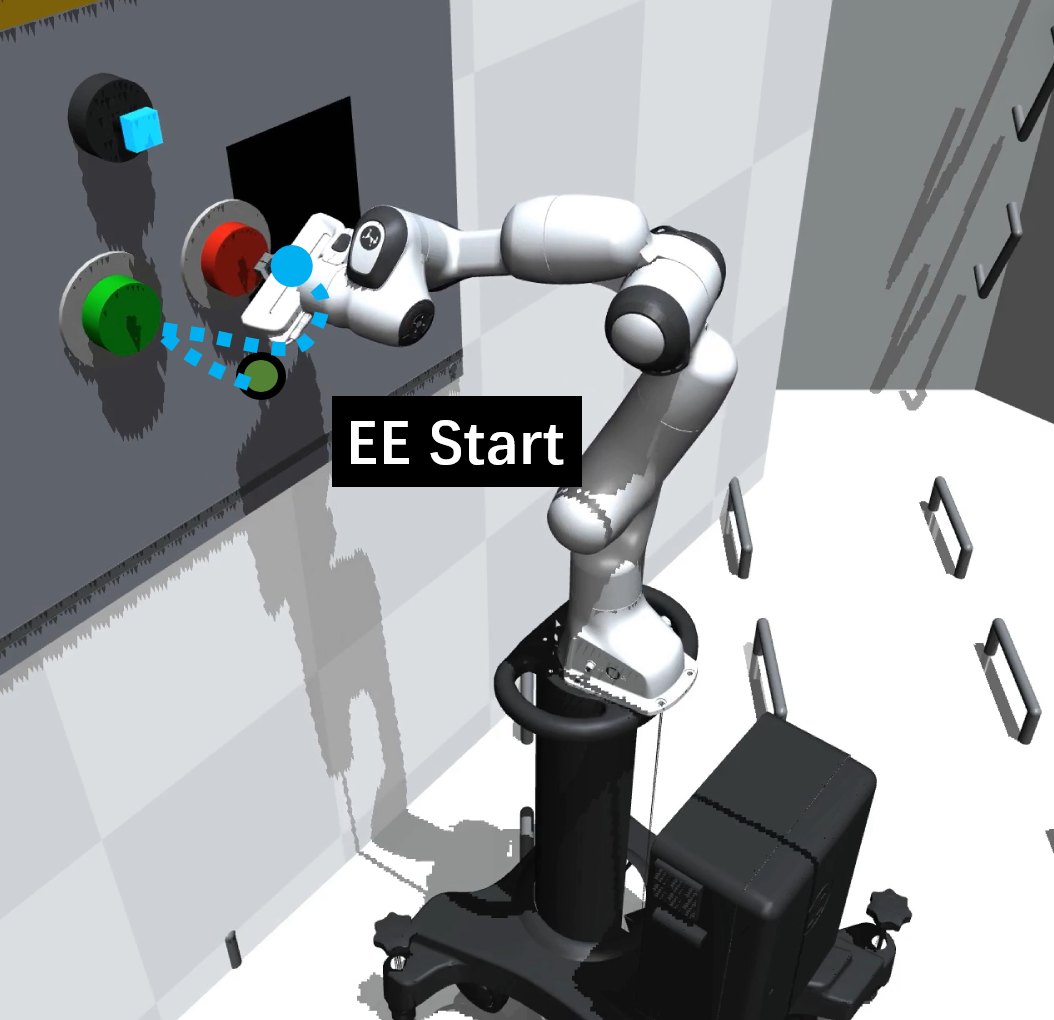} \hfill
		\includegraphics[width=0.23\linewidth]{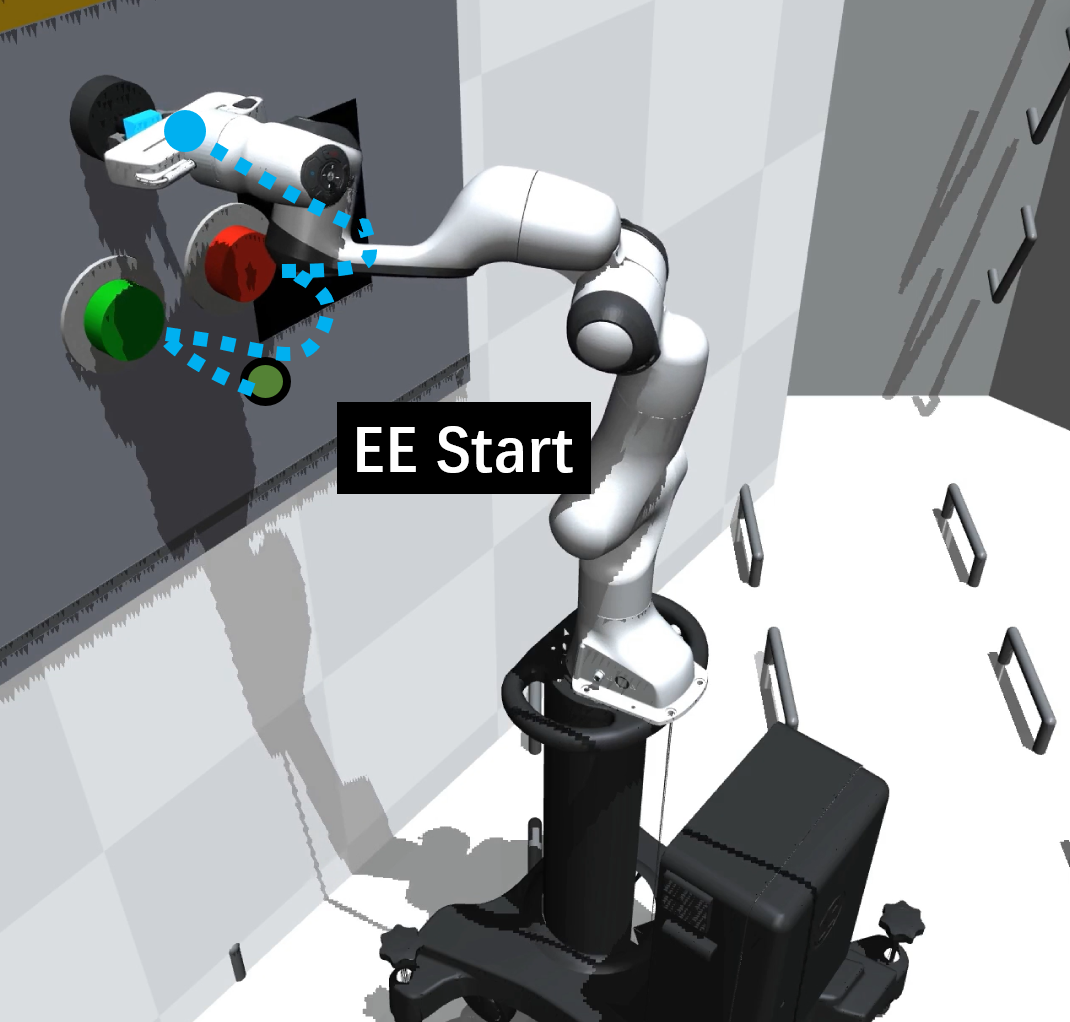} \hfill
		\includegraphics[width=0.23\linewidth]{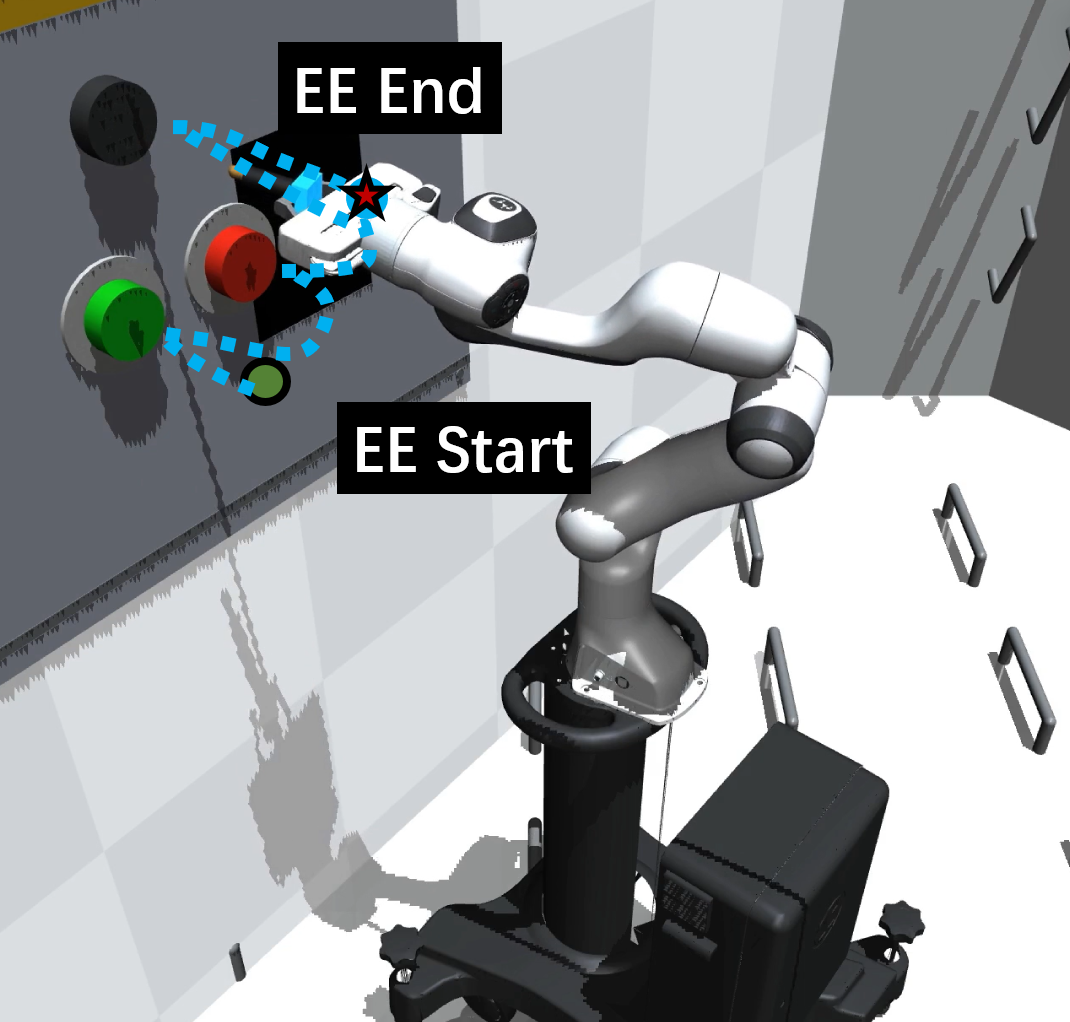}
	}
	
	\subfloat[Task V: Long-horizon Tumbling Target Stowing in a Drawer]{
		\includegraphics[width=0.23\linewidth]{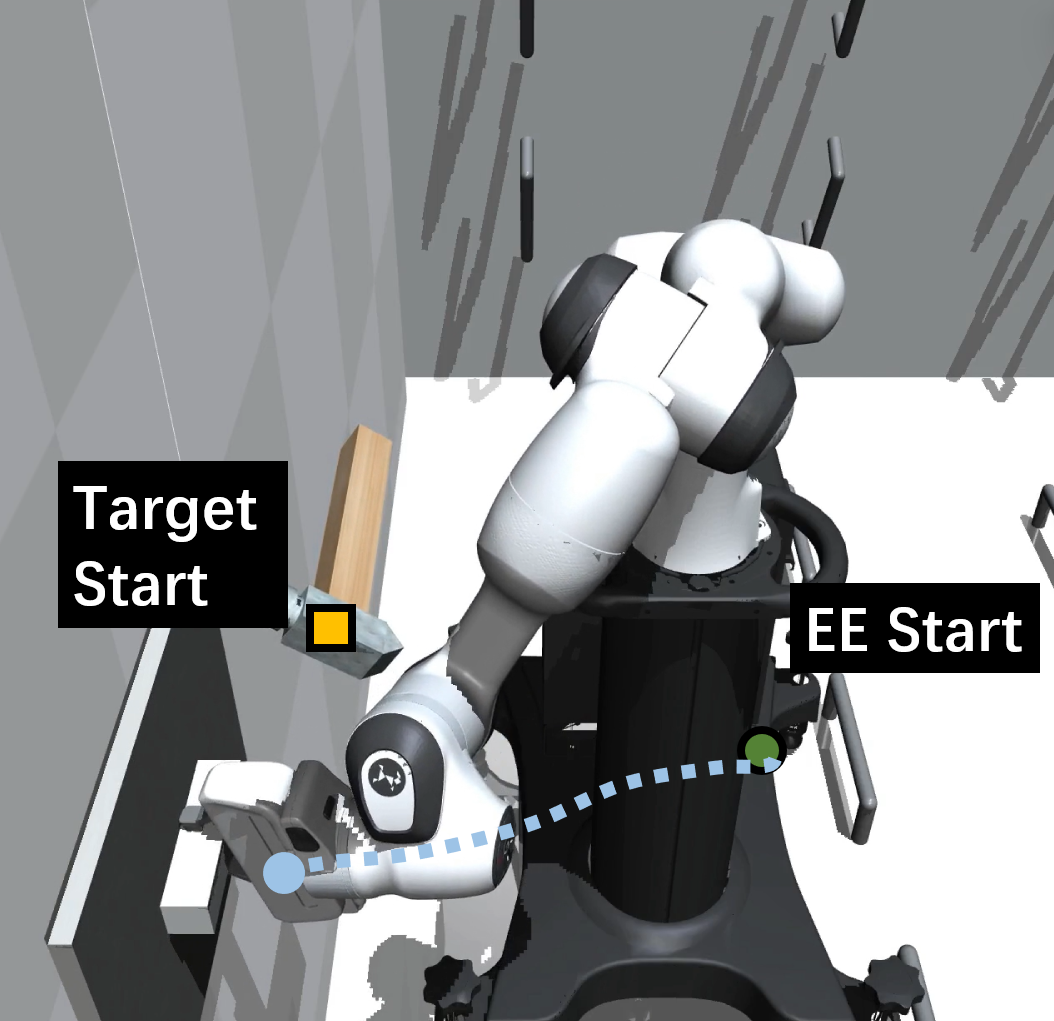} \hfill
		\includegraphics[width=0.23\linewidth]{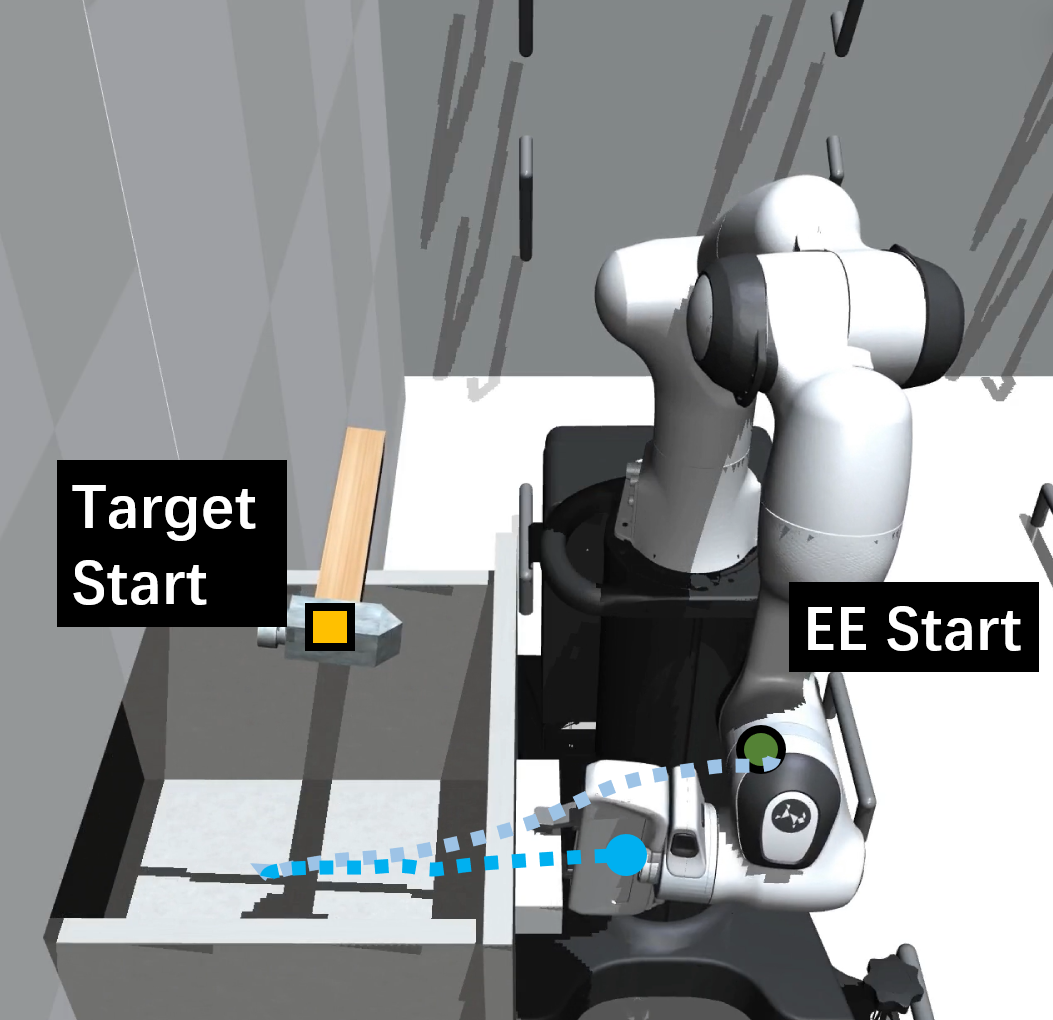} \hfill
		\includegraphics[width=0.23\linewidth]{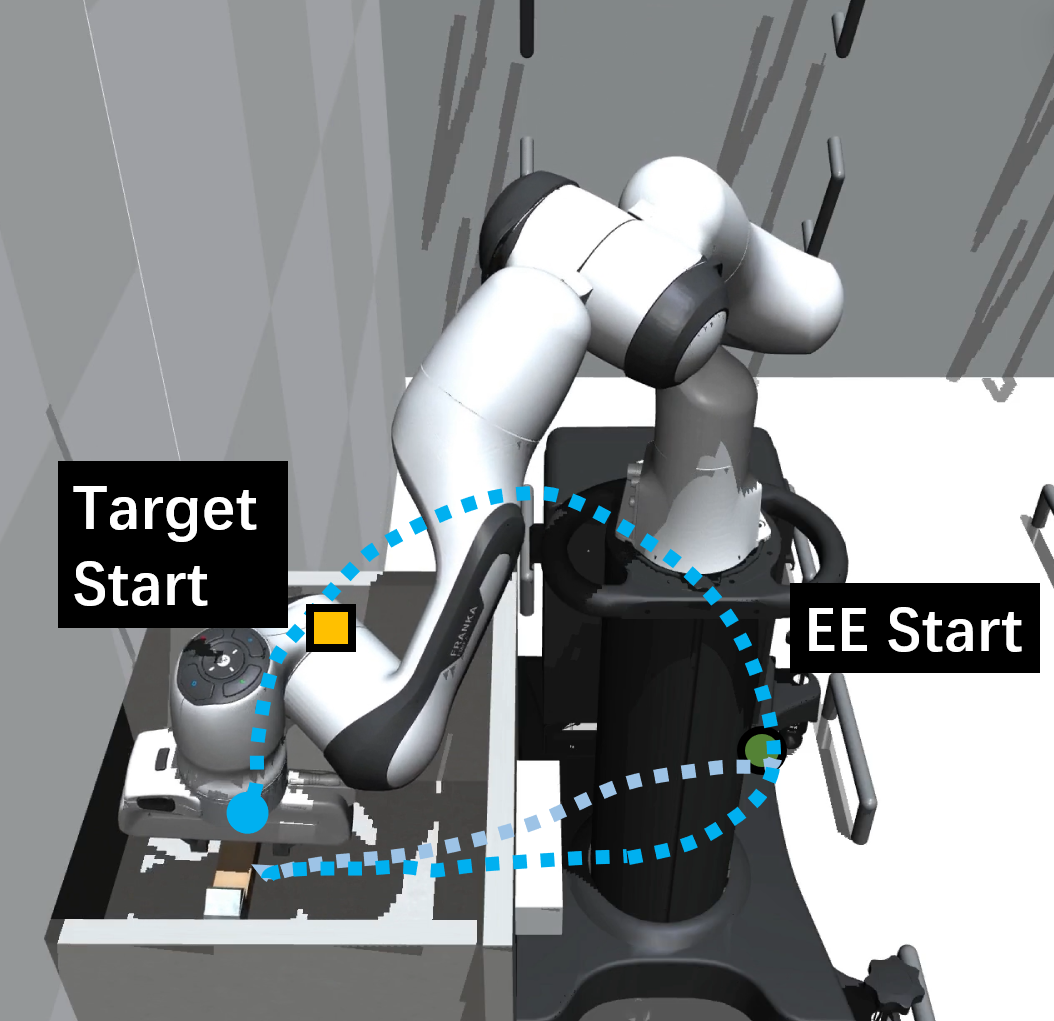} \hfill
		\includegraphics[width=0.23\linewidth]{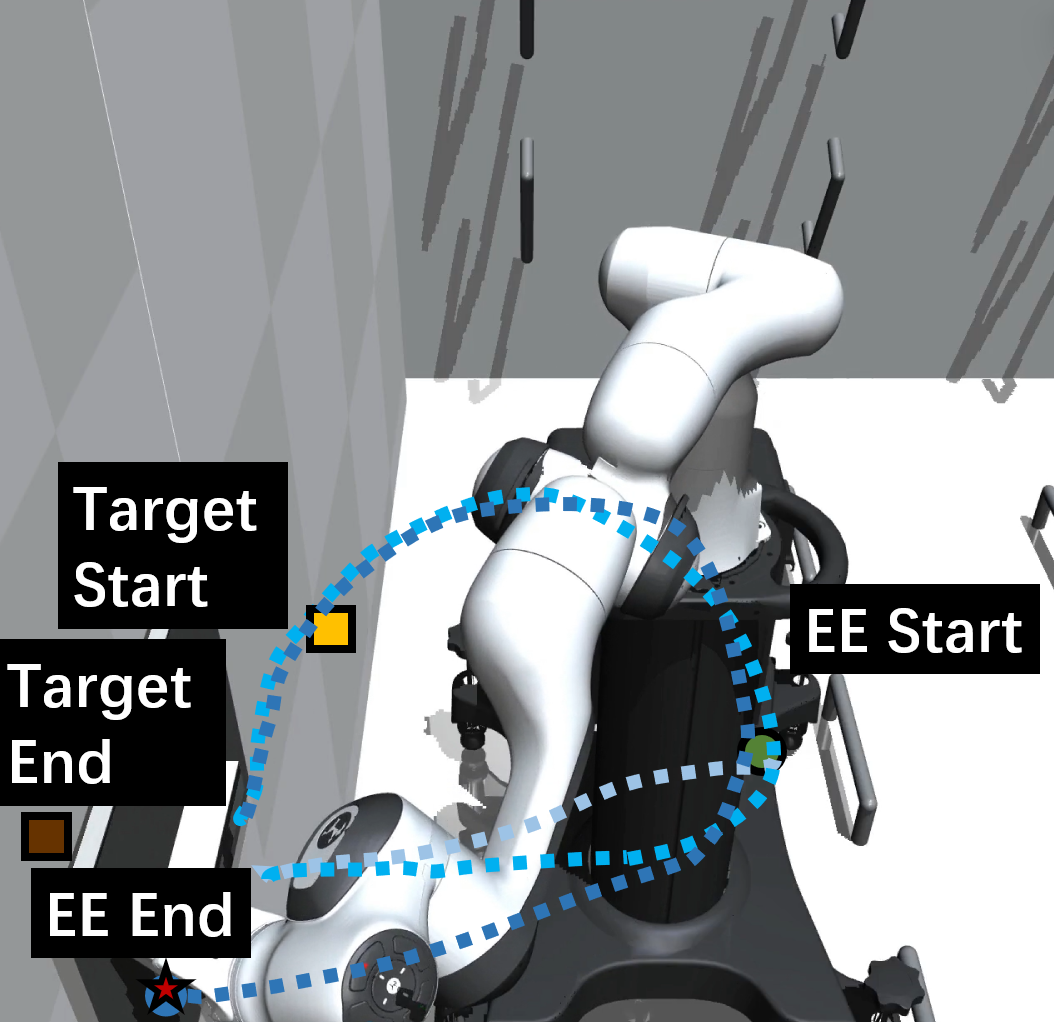}
	}
	
	\caption{Five representative images of the experimental environment for space capsule operation tasks.}
	\label{fig:env_images}
\end{figure}

Recent studies have turned to diffusion policies (DP) \cite{chi2025diffusion} to solve these issues\cite{briden2025diffusion}. By modeling robot behavior as a conditional denoising process, DP captures complex distributions while maintaining execution stability. However, the iterative sampling in diffusion models is computationally expensive. This high energy demand often exceeds the strict power limits of robots operating on resource-constrained space platforms.

To bridge this gap, we leverage the event-driven nature of Spiking Neural Networks (SNNs) to achieve low-power inference \cite{tavanaei2019deep, eshraghian2023training, zhang2026multimodal}. This paper proposes Spiking Diffusion Policies (SDP), a generative approach designed for precise control under strict power limits. Building on this concept, we propose L-SDPPO, an SNN-based RL framework for fine-tuning SDP. By integrating the SDLI mechanism to refine spatiotemporal perception, L-SDPPO significantly enhances the adaptability of spiking policies in microgravity. By mimicking biological neural delays to regulate input timing, SDLI improves the perception of temporal features during manipulation. Our framework maintains high success rates while significantly reducing energy use during microgravity operations.

The main contributions of this work are summarized as follows:

1. We present the L-SDPPO framework for intra-vehicular robotic manipulations, which integrates an SDLI mechanism to augment spatiotemporal sensitivity. This approach refines the discernment of dynamic temporal cues, thereby facilitating energy-efficient and precise onboard inference.

2. A benchmark for intra-vehicular robotic manipulation is presented, in which an expert dataset is curated across five representative scenarios. This benchmark covers high-precision tasks such as hatch operation and dynamic stowing to address the inherent challenges of manipulation in microgravity.

3. Experimental results demonstrate that L-SDPPO maintains superior task success rates under microgravity disturbances while substantially reducing power consumption.

\section{Related Work}

\subsection{Space Robotic Manipulations}
Model-based impedance control was the primary method in early research on precision space robot operations \cite{wang2014emg, platt2011multiple}. For instance, Palma et al. \cite{palma2023application} used impedance control for debris capture with free-floating space manipulators. The robot tunes its dynamics to safely interact with tumbling targets. Liu et al. \cite{liu2019research} proposed a position-based Cartesian impedance control strategy for on-orbit screwing missions. Their approach utilizes joint friction identification to resolve the conflict between high positioning accuracy and the compliant force requirements of flexible joints. However, these traditional strategies often require complex parameter tuning for specific mechanical structures, making them difficult to generalize to unstructured tasks.

RL has been widely explored to automate motion planning for space manipulators \cite{peng2024reinforcement, hu2025deep}. Specifically, Li et al. \cite{li2021constrained} utilized the Deep Deterministic Policy Gradient (DDPG) algorithm for space dual-arm manipulator to handle self-collision and velocity limits. Other DDPG-based frameworks have addressed path planning for 6-DOF manipulators and collision avoidance for irregular targets \cite{al2024path, blaise2023space}. Despite these successes, RL methods often suffer from low sample efficiency, making them difficult to train in space environments with sparse rewards \cite{nair2018overcoming, vecerik2017leveraging}.

To address these challenges of low sample efficiency and sparse rewards, researchers have increasingly turned to IL, which leverages expert demonstrations to rapidly acquire near-optimal robotic manipulation policies\cite{shao2026imitation}. Ashith Shyam et al. \cite{shyam2020imitation, ashith2021autonomous} used expert trajectories from model predictive control to train Probabilistic Movement Primitives for 7-DOF space manipulators. This approach enables high-precision motion through fast probabilistic inference. It also helps mitigate reaction forces to maintain spacecraft attitude stability. Joukov and Kuli\'{c} \cite{joukov2017gaussian} applied Gaussian Processes to learn task-specific cost functions. These functions were integrated into linearized model predictive control to meet joint and task space constraints. Ning et al. \cite{ning2025integrated} developed the GD-IL framework for space robotics. This method uses second-order dynamic movement primitives to achieve robust grasping and avoid singularities. These implementations successfully reproduce trajectories. However, they rely on unimodal representations. This limitation prevents them from modeling the multimodal action distributions found in dynamic space environments. Our L-SDPPO framework addresses these representation constraints by utilizing SDP to capture complex multimodal distributions accurately. By leveraging event-driven neural dynamics, our method provides the necessary precision for dynamic manipulation while substantially reducing the inference energy consumption required by traditional methods.

\subsection{DP-based Robotic Manipulations}

Recent research uses diffusion-based generative modeling to overcome previous limitations. Janner et al. \cite{janner2022planning} introduced the Diffuser framework to represent robot trajectories as samples. Chi et al. \cite{chi2025diffusion} applied this to high-precision policy learning. Their work shows that predicting action sequences captures the multimodal distributions of human demonstrations. Other studies scale these policies to more diverse tasks. Ze et al. \cite{ze20243d} used 3D visual data to improve spatial robustness. The Octo model \cite{team2024octo} employs transformers and large-scale pretraining for generalist policies. But increased model complexity leads to high computational costs and latency. Researchers have proposed hierarchical methods \cite{ma2024hierarchical} or discrete formulations \cite{yu2025d3p} to improve efficiency. 

Despite the ability of traditional diffusion models to capture complex action distributions, their effectiveness remains limited by the quality of expert demonstrations. To move beyond the constraints of datasets and achieve super-expert performance, Diffusion Policy Policy Optimization (DPPO) employs RL to fine-tune pre-trained policies \cite{ren2024diffusion}. Despite high performance, these diffusion models demand significant computational power, making deployment difficult on space platforms with low energy budgets. Our work addresses this gap by introducing SDP, a neuromorphic generative framework that parameterizes the iterative denoising process through a fully spiking residual architecture. By leveraging the event-driven nature and temporal sparsity of spiking neurons, SDP is specifically designed to achieve the high modeling precision of DP while strictly adhering to the energy constraints of onboard space missions. By integrating SDP with the SDLI mechanism, our approach achieves exceptional precision and reduced energy consumption during fine manipulation in microgravity.

\section{Methodology}
In this section, we present the L-SDPPO framework. We first formulate the SDP architecture, which parameterizes the generative denoising process using a fully spiking residual network. Subsequently, we introduce the SDLI mechanism and optimize the policy via PPO.

\begin{figure*}
	\centering
	\includegraphics[width=\linewidth]{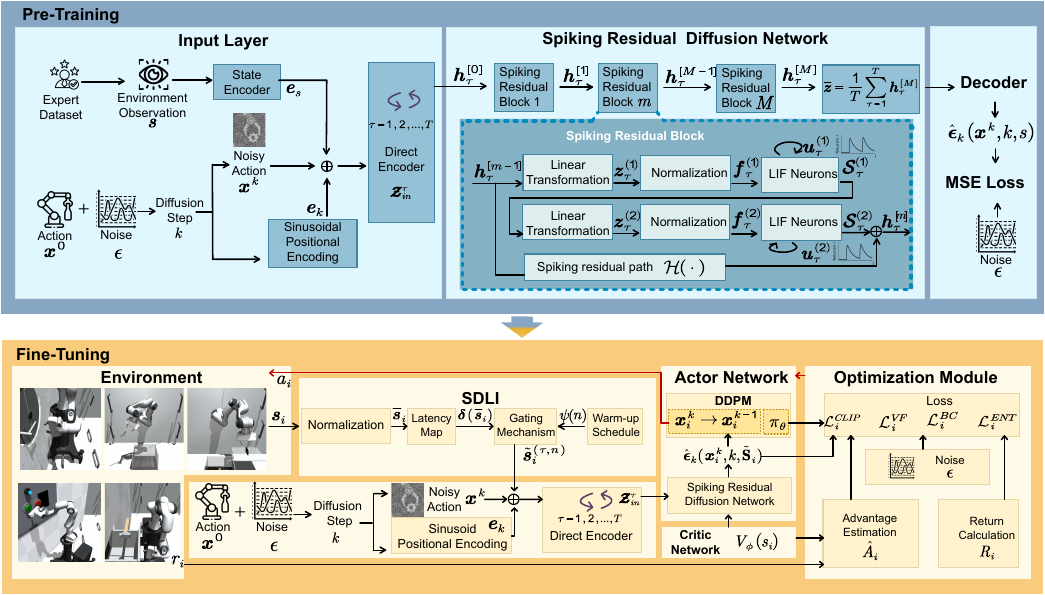}
	\caption{Overview of the L-SDPPO framework. The framework consists of two phases: (Top) Pre-Training, where the spiking residual diffusion network is trained to predict noise $\hat{\epsilon}_k$ using an expert dataset via MSE loss; (Bottom) Fine-Tuning, which incorporates the SDLI module to modulate environmental state features before online optimization via a multi-component PPO loss.}
	\label{fig:placeholder}
\end{figure*}

\subsection{SDP}

Policy generation is based on a conditional Denoising Diffusion Probabilistic Model (DDPM). Following the formulation
in \cite{chi2025diffusion}, we define the policy $\pi_\theta$ as a $K$-step reverse denoising process conditioned on the environment observation $\boldsymbol{s}$. The policy generates physical actions by iteratively denoising Gaussian noise $\boldsymbol{x}^K \sim \mathcal{N}(\mathbf{0}, \mathbf{I})$. For each denoising step $k \in \{K, K-1, \dots, 1\}$, the state transition is modeled as:
\begin{equation}
	\boldsymbol{x}^{k-1} \!\sim\! p_\theta(\boldsymbol{x}^{k-1} \mid \boldsymbol{x}^k, \boldsymbol{s}) = \mathcal{N}(\boldsymbol{x}^{k-1}; \boldsymbol{\mu}_\theta(\boldsymbol{x}^k, k, \boldsymbol{s}), \sigma_k^2 \mathbf{I}),
\end{equation}
where $p_\theta$ denotes the distribution of the reverse diffusion chain, the mean $\boldsymbol{\mu}_\theta$ is parameterized by a neural network, and $\sigma_k$ follows a predefined variance schedule. The final denoised output $\boldsymbol{x}^0$ is utilized as the physical action $\boldsymbol{a}$ for environment interaction. 

To achieve high energy efficiency, we parameterize the noise prediction network $\boldsymbol{\mu}_\theta$ using a SNN. On the macroscale, the diffusion denoising sequence (step $k$) governs the generation of physical actions, where each denoising step invokes a complete forward pass of the SNN backbone. On the microscale, the SNN dynamics evolve over a time window $\tau \in \{1, \dots, T\}$. As shown in Fig. \ref{fig:placeholder}, this SDP architecture integrates hybrid input encoding, Leaky Integrate-and-Fire (LIF) dynamics, and full-spiking residual blocks.

\subsubsection{Hybrid Input Encoding}
To process multi-modal inputs, a unified encoding module maps continuous sensory data into the spiking domain. The discrete timestep $k$ is embedded into a temporal feature $\boldsymbol{e}_k$ via sinusoidal positional encoding, while the observation $\boldsymbol{s}$ is processed by a state encoder to extract the conditional feature $\boldsymbol{e}_s$. These are concatenated with the noisy action $\boldsymbol{x}^k$ to form the joint input $\boldsymbol{\mathcal{Z}}_{in} \in \mathbb{R}^{D_{in}}$:
\begin{equation}
	\boldsymbol{\mathcal{Z}}_{in} = [\boldsymbol{x}^k, \boldsymbol{e}_k, \boldsymbol{e}_s].
\end{equation}

Following the direct coding paradigm, $\boldsymbol{\mathcal{Z}}_{in}$ is injected as a constant current into the SNN across all internal simulation timesteps $\tau$:
\begin{equation}
	\boldsymbol{h}_\tau^{[0]} = \boldsymbol{\mathcal{Z}}_{in}, \quad \forall \tau \in \{1, \dots, T\}.
\end{equation}

\subsubsection{Spatiotemporal Forward Propagation}
Within the microscale time window $T$, signals propagate through $M$ full-spiking residual blocks. The joint update of the block output $\boldsymbol{h}_\tau^{[m]}$ and membrane potential $\boldsymbol{u}_\tau^{[m]}$ is defined as $(\boldsymbol{h}_\tau^{[m]}, \boldsymbol{u}_\tau^{[m]}) = \mathcal{F}_{\text{block}}^{[m]}(\boldsymbol{h}_\tau^{[m-1]}, \boldsymbol{u}_{\tau-1}^{[m]})$.

Given the spiking input $\boldsymbol{h}_\tau^{[m-1]}$, the core spike transformation integrates linear transformations, numerical normalization, and LIF neuronal dynamics. The normalized current $\boldsymbol{f}_\tau^{(1)}$ is computed as:
\begin{equation}
	\boldsymbol{z}_\tau^{(1)} = \mathbf{W}_1 \boldsymbol{h}_\tau^{[m-1]} + \boldsymbol{b}_1,
\end{equation}
\begin{equation}
	\boldsymbol{f}_\tau^{(1)} = \boldsymbol{\gamma}_1 \odot \frac{\boldsymbol{z}_\tau^{(1)} - \mathbb{E}[\boldsymbol{z}_\tau^{(1)}]}{\sqrt{\text{Var}[\boldsymbol{z}_\tau^{(1)}] + \varrho}} + \boldsymbol{{\vartheta}_1},
\end{equation}
where $\{\mathbf{W}_1, \boldsymbol{b}_1\}$ are learnable weights and biases, $\{\boldsymbol{\gamma}_1, \boldsymbol{\vartheta}_1\}$ are learnable parameters, $\mathbb{E}[\cdot]$ and $\text{Var}[\cdot]$ are the mean and variance across feature dimensions, $\odot$ is the Hadamard product operator indicating element-wise multiplication, and $\varrho$ is a small constant added to the variance to prevent division by zero. 

The current $\boldsymbol{f}_\tau^{(1)}$ is then integrated into the LIF membrane potential $\tilde{\mathbf{u}}_\tau^{(1)}$ with a decay factor $\beta \in (0,1)$:
\begin{equation}
	\tilde{\mathbf{u}}_\tau^{(1)} = \beta \mathbf{u}_{\tau-1}^{(1)} + \boldsymbol{f}_\tau^{(1)}.
\end{equation}

A spike $\boldsymbol{\mathcal{S}}_\tau^{(1)}$ is generated via the Heaviside step function $\Theta(\cdot)$ when the potential exceeds the threshold $u_{\text{thr}}$, followed by a soft reset:
\begin{equation}
	\boldsymbol{\mathcal{S}}_\tau^{(1)} = \Theta(\tilde{\mathbf{u}}_\tau^{(1)} - u_{\text{thr}}),
\end{equation}
\begin{equation}
	\mathbf{u}_\tau^{(1)} = \tilde{\mathbf{u}}_\tau^{(1)} - \boldsymbol{\mathcal{S}}_\tau^{(1)} \cdot u_{\text{thr}}.
\end{equation}

To enable backpropagation, the non-differentiable spike generation is approximated using a surrogate gradient:
\begin{equation}
	\frac{\partial \boldsymbol{\mathcal{S}}}{\partial \mathbf{u}} \approx (1 + g|\mathbf{u} - u_{\text{thr}}|)^{-2},
\end{equation}
where $g$ controls the steepness of the surrogate function.

After cascading a secondary LIF layer to produce $\boldsymbol{\mathcal{S}}_\tau^{(2)}$, residual fusion is performed using sparse accumulate (AC) operations:
\begin{equation}
	\boldsymbol{h}_\tau^{[m]} = \boldsymbol{\mathcal{S}}_\tau^{(2)} + \mathcal{H}(\boldsymbol{h}_\tau^{[m-1]}),
\end{equation}
where $\mathcal{H}(\cdot)$ is the spiking residual path. This fused multi-level spiking vector is passed to block $m+1$.

Finally, global temporal decoding is applied to compute a global feature vector $\bar{\boldsymbol{z}}$ by averaging the temporal sequence from the final block $M$:
\begin{equation}
	\bar{\boldsymbol{z}} = \frac{1}{T} \sum_{\tau=1}^{T} \boldsymbol{h}_\tau^{[M]}.
\end{equation}

The averaged representation $\bar{\boldsymbol{z}}$ is then decoded to produce the final continuous noise prediction $\hat{\boldsymbol{\epsilon}}_k$, which subsequently parameterizes the distribution mean $\boldsymbol{\mu}_\theta$ for the current diffusion step $k$.

\subsubsection{Training Objective}
The SDP is trained via supervised learning on an expert dataset $\mathcal{D}$. During the pre-training phase, the network learns to reverse the forward diffusion process by predicting the added noise. 

For each training iteration, we sample an expert action $\boldsymbol{x}^0$ and a random diffusion timestep $k$. The corresponding noisy action $\boldsymbol{x}^k$ is constructed using the forward process:
\begin{equation}
	\boldsymbol{x}^k = \sqrt{\bar{\alpha}_k} \boldsymbol{x}^0 + \sqrt{1 - \bar{\alpha}_k} \boldsymbol{\epsilon}, \quad \boldsymbol{\epsilon} \sim \mathcal{N}(\mathbf{0}, \mathbf{I}),
\end{equation}
where $\bar{\alpha}_k$ denotes the cumulative product of the variance schedule at step $k$.

The objective of the optimization is to minimize the discrepancy between the target noise and the prediction from the SNN backbone. The pre-training loss $\mathcal{L}_{\text{pre}}$ is formulated as the Mean Squared Error (MSE):
\begin{equation}
	\mathcal{L}_{\text{pre}} = \mathbb{E}_{k, \boldsymbol{x}^0, \boldsymbol{\epsilon}} \left[ \left\| \boldsymbol{\epsilon} - \hat{\boldsymbol{\epsilon}}_k(\boldsymbol{x}^k, k, \boldsymbol{s}) \right\|^2 \right].
\end{equation}
\subsection{SDLI}
We use a constant input current $\boldsymbol{\mathcal{Z}}_{in}$ during pre-training. For fine-tuning, we introduce SDLI to control the timing of state features entering the spiking network (Fig. \ref{fig:placeholder}). This mechanism follows the biological time-to-first-spike (TTFS) coding principle. In TTFS, higher stimulus intensity leads to shorter response latencies. Similarly, SDLI maps state magnitudes to specific temporal delays. This modulation improves the spatiotemporal sensitivity of the model. It allows spiking neurons to capture the underlying dynamics necessary for environment interaction.

\paragraph{Latency Mapping}
For a raw state vector $\boldsymbol{s}_i$ at environment step $i$, we first compute the normalized intensity $\bar{\boldsymbol{s}}_i$:
\begin{equation}
	\bar{\boldsymbol{s}}_i = \frac{\boldsymbol{s}_i - \min(\boldsymbol{s}_i)}{\max(\boldsymbol{s}_i) - \min(\boldsymbol{s}_i) + \eta},
\end{equation}\label{eq14}
where $\eta$ is a small constant to prevent division by zero. 

We denote the resulting delay vector as $\boldsymbol{\delta}(\bar{\boldsymbol{s}}_i)$:
\begin{equation}
	\boldsymbol{\delta}(\bar{\boldsymbol{s}}_i) = T_{\min} + (T_{\max} - T_{\min}) \cdot (1 - \bar{\boldsymbol{s}}_i),
\end{equation}\label{eq15}
where $[T_{\min}, T_{\max}] \subset [0, T]$ specifies the allowable delay interval within the SNN time window.

\paragraph{Soft Gating Mechanism for Latency Injection}
We formulate the latency injection as a soft gate to facilitate gradient propagation. At training iteration $n$, the effective state input $\tilde{\boldsymbol{s}}_i^{(\tau, n)}$ received by the SNN at internal timestep $\tau$ is defined as:
\begin{equation}
	\tilde{\boldsymbol{s}}_i^{(\tau, n)}= {\boldsymbol{s}_i} \odot \left[ (1 - \psi(n)) + \psi(n) \cdot \sigma \left( \frac{\tau - \boldsymbol{\delta}(\bar{\boldsymbol{s}}_i)}{\varkappa} \right) \right],
\end{equation}
where $\sigma(\cdot)$ denotes the sigmoid activation and $\varkappa$ governs the gate steepness. To ensure stable convergence, the injection strength $\psi(n)$ is modulated via a linear warm-up schedule:
\begin{equation}
	\psi(n) = \psi_{\max} \cdot \min \left( 1, \frac{n}{N_w} \right),
\end{equation}
with $n$, $N_w$, and $\psi_{\max}$ representing the current iteration, warm-up duration, and peak strength, respectively.

\subsection{PPO-based SDP Optimization}
To address the robustness deficiencies of IL in microgravity resulting from limited expert coverage and cumulative errors, we employ PPO to fine-tune the SDP online. This active exploration mechanism significantly enhances the robot?s dynamic control capabilities against unconstrained object drift.

The time-varying sequence $\tilde{\mathbf{S}}_i = \big( \tilde{\boldsymbol{s}}_i^{{(\tau, n)}} \big)_{\tau=1}^T$ provides the dynamic conditioning for the spiking backbone. The resulting joint distribution of the reverse diffusion trajectory is formulated as:
\begin{equation}
	\pi_\theta(\boldsymbol{x}_i^{0:K} \mid \tilde{\mathbf{S}}_i) = p(\boldsymbol{x}_i^K) \prod_{k=1}^K \pi_\theta(\boldsymbol{x}_i^{k-1} \mid \boldsymbol{x}_i^k, k, \tilde{\mathbf{S}}_i),
\end{equation}
where the initial noise follows $\boldsymbol{x}_i^K \sim \mathcal{N}(\mathbf{0}, \mathbf{I})$, and each denoising transition is modeled by a Gaussian distribution:
\begin{equation}
	\pi_\theta(\boldsymbol{x}_i^{k-1} \mid \boldsymbol{x}_i^k, k, \tilde{\mathbf{S}}_i) = \mathcal{N}(\boldsymbol{x}_i^{k-1}; \boldsymbol{\mu}_\theta(\boldsymbol{x}_i^k, k, \tilde{\mathbf{S}}_i), \sigma_k^2 \mathbf{I}).
\end{equation}

We optimize the policy by minimizing the total objective function via gradient descent:
\begin{equation}
	\mathcal{L}_{\text{total}} = \frac{1}{N} \sum_{i=1}^{N} \left( \mathcal{L}_i^{CLIP} \!+\!c_{v} \mathcal{L}_i^{VF} \!+\! c_{b} \mathcal{L}_i^{BC} \!+\! c_{e} \mathcal{L}_i^{ENT} \right),
\end{equation}
where the hyperparameters $c_{v}$, $c_{b}$, and $c_{e}$ denote the trade-off weights for the value loss, behavior cloning (BC) loss, and entropy regularization, respectively.

The policy update is constrained by the PPO clipping objective to prevent destabilizing updates. For a given denoising step $k$:
\begin{equation}
	\mathcal{L}_i^{CLIP} \!=\! -\min (p_{r_i}(\theta) \hat{A}_i, \text{clip}(p_{r_i}(\theta), 1\!-\!\xi, 1\!+\!\xi) \hat{A}_i),
\end{equation}
where the probability ratio is $p_{r_i}(\theta) = \frac{\pi_\theta(\boldsymbol{x}_i^{k-1} \mid \boldsymbol{x}_i^k, k, \tilde{\mathbf{S}}_i)}{\pi_{\text{old}}(\boldsymbol{x}_i^{k-1} \mid \boldsymbol{x}_i^k, k, \tilde{\mathbf{S}}_i)}$, $\hat{A}_i$ denotes the advantage estimate, and $\xi$ is the clipping hyperparameter.

Simultaneously, the value function $V_{\phi}(\boldsymbol{s}_i)$ is optimized by minimizing the MSE against the target return $R_i$:

\begin{equation}
	\mathcal{L}_i^{VF} = \frac{1}{2} (V_\phi(\boldsymbol{s}_i) - R_i)^2,
\end{equation}
where the target return is defined as $R_i = \hat{A}_i + V_\phi(\boldsymbol{s}_i)$.

To maintain consistency with the expert distribution from the pre-training phase, a BC loss is incorporated to penalize the deviation of the spiking network's output from the expert noise:
\begin{equation}
	\mathcal{L}_i^{BC} = \mathbb{E}_{k, \boldsymbol{x}_i^0, \boldsymbol{\epsilon}} \left[ \| \boldsymbol{\epsilon} - \hat{\boldsymbol{\epsilon}}_k (\boldsymbol{x}_i^k, k, \tilde{\mathbf{S}}_i) \|^2 \right],
\end{equation}
where $\boldsymbol{\epsilon} \sim \mathcal{N}(\mathbf{0}, \mathbf{I})$ is the ground-truth noise, and $\hat{\boldsymbol{\epsilon}}_k$ is the noise prediction generated by the SNN backbone after $T$ internal timesteps.

To promote exploration and avoid local optima, an entropy regularization term is added for each sample $i$:
\begin{equation}
	\mathcal{L}_i^{ENT} = -\omega_i,
\end{equation}
where $\omega_i$ represents the empirical variance of the sampled action distribution.

\section{EXPERIMENTS}

This section evaluates the L-SDPPO framework in simulated intra-cabin tasks to verify its performance superiority, energy efficiency, and component-wise validity. We first detail the experimental setups (Sec.~\ref{subsec:experiment_settings}). Next, we provide a qualitative analysis of the robot's execution trajectories across all tasks (Sec.~\ref{subsec:qualitative_analysis}). We then introduce the energy consumption model (Sec.~\ref{subsec:energy_model}) and present a quantitative comparative performance analysis (Sec.~\ref{subsec:performance_eval}), finally concluding with ablation studies (Sec.~\ref{subsec:ablation_study}).

\subsection{Experiment Settings}
\label{subsec:experiment_settings}
To evaluate the proposed L-SDPPO framework, we benchmark it against several methods, including Gaussian-based Behavior Cloning (GBC) \cite{chernova2007confidence}, PPO \cite{schulman2017proximal}, GBC pre-training with PPO fine-tuning (GPPO) \cite{nair2018overcoming}, DP \cite{chi2025diffusion}, and DPPO \cite{ren2024diffusion}. Performance is quantified using three primary evaluation metrics: Success Rate (SR) to measure task completion capability, Average Reward (AR) to assess policy effectiveness, and Energy Consumption (EC) to evaluate the energy efficiency of the policies.

The training for L-SDPPO operates in two phases. During the pre-training phase, the SDP is training via IL using 200 expert demonstration trajectories collected for each specific task. The base policy is optimized over 3,000 epochs to yield a robust prior distribution for robotic actions. In the subsequent post-training phase, the policy undergoes online refinement via PPO. The RL fine-tuning leverages 50 parallel environments to collect experience, sampling a batch size of 10,000 transitions per update cycle, and optimizes the surrogate objective over 10 epochs. 

To demonstrate the generalizability of our approach, we deploy a unified set of hyperparameters across all tasks. The core hyperparameters of the proposed method are summarized in Table~\ref{tab:hyperparameters}.

\begin{table}[htbp]
	\centering
	\caption{Core Hyperparameters for L-SDPPO}
	\label{tab:hyperparameters}
	\footnotesize 
	\renewcommand{\arraystretch}{1.2} 
	\setlength{\tabcolsep}{0pt} 
	
	\begin{tabular*}{\columnwidth}{@{\extracolsep{\fill}}lc | lc@{}}
		\toprule
		\textbf{Parameter} & \textbf{Value} & \textbf{Parameter} & \textbf{Value} \\ \midrule
		\multicolumn{4}{c}{\textit{Diffusion \& Architecture Settings}} \\ \midrule
		Obs. / Action Horizon & 1 / 4 & Actor Hidden Dims & [2048, 2048] \\
		Pretrain Denoising Steps & 20 & Critic Hidden Dims & [256, 256, 256] \\
		Finetune Denoising Steps & 10 & Actor / Critic LR & $10^{-4}$ / $10^{-3}$ \\
		\midrule
		\multicolumn{4}{c}{\textit{PPO Fine-tuning Settings}} \\ \midrule
		Discount Factor  & 0.999 & GAE Parameter  & 0.95 \\
		PPO Batch Size & 10000 & Update Epochs & 10 \\
		\midrule
		\multicolumn{4}{c}{\textit{SNN \& SDLI Mechanism Settings}} \\ \midrule
		SNN Time Steps  & 3 & Latency Bounds  & 1.0, 3.0 \\
		Max SDLI Strength  & 0.2 & SDLI Smoothness  & 0.9 \\
		\bottomrule
	\end{tabular*}
\end{table}

\begin{figure*}[htbp]
	\centering
	\captionsetup[subfloat]{font={rm,small}, labelformat=simple}
	\renewcommand{\thesubfigure}{(\alph{subfigure})}
	\newcounter{tempcounter}
	\setcounter{tempcounter}{1} 
	\subfloat[Task Trajectories.]{
		\begin{minipage}{0.2\linewidth}
			\centering
			\includegraphics[width=\linewidth]{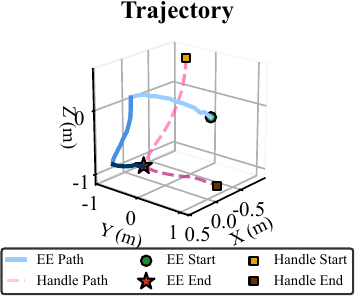} \\
			\vspace{1mm} \small (\arabic{tempcounter}) Task I \stepcounter{tempcounter}
		\end{minipage}\hfill
		\begin{minipage}{0.2\linewidth}
			\centering
			\includegraphics[width=\linewidth]{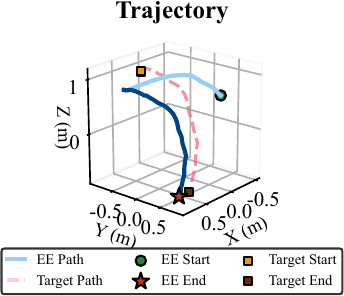} \\
			\vspace{1mm} \small (\arabic{tempcounter}) Task II \stepcounter{tempcounter}
		\end{minipage}\hfill
		\begin{minipage}{0.18\linewidth}
			\centering
			\includegraphics[width=\linewidth]{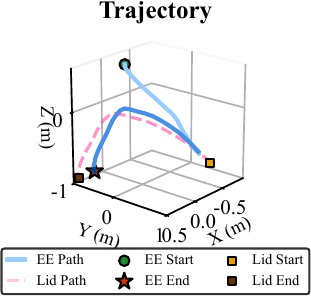} \\
			\vspace{1mm} \small (\arabic{tempcounter}) Task III \stepcounter{tempcounter}
		\end{minipage}\hfill
		\begin{minipage}{0.18\linewidth}
			\centering
			\includegraphics[width=\linewidth]{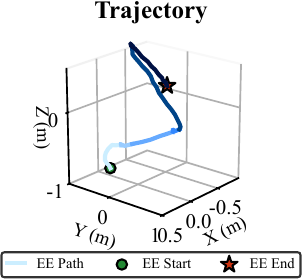} \\
			\vspace{1mm} \small (\arabic{tempcounter}) Task IV \stepcounter{tempcounter}
		\end{minipage}\hfill
		\begin{minipage}{0.2\linewidth}
			\centering
			\includegraphics[width=\linewidth]{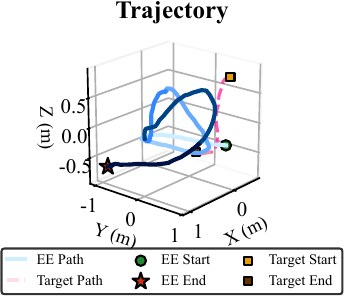} \\
			\vspace{1mm} \small (\arabic{tempcounter}) Task V
		\end{minipage}
	}
	\vspace{4mm} 
	\setcounter{tempcounter}{1} 
	\subfloat[Stage-wise State Transitions.]{
		\begin{minipage}{0.18\linewidth}
			\centering
			\includegraphics[width=\linewidth]{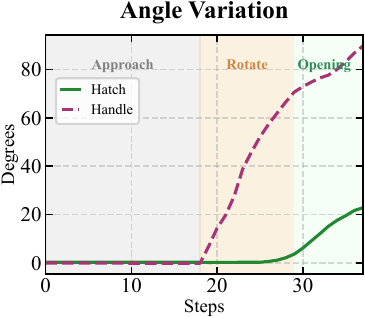} \\
			\vspace{1mm} \small (\arabic{tempcounter}) Task I \stepcounter{tempcounter}
		\end{minipage}\hfill
		\begin{minipage}{0.18\linewidth}
			\centering
			\includegraphics[width=\linewidth]{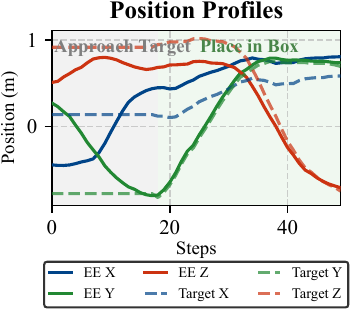} \\
			\vspace{1mm} \small (\arabic{tempcounter}) Task II \stepcounter{tempcounter}
		\end{minipage}\hfill
		\begin{minipage}{0.18\linewidth}
			\centering
			\includegraphics[width=\linewidth]{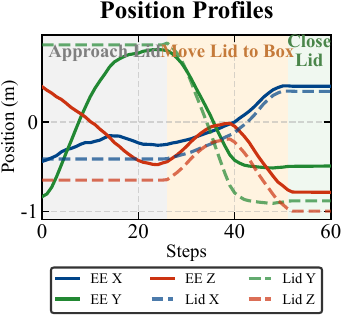} \\
			\vspace{1mm} \small (\arabic{tempcounter}) Task III \stepcounter{tempcounter}
		\end{minipage}\hfill
		\begin{minipage}{0.2\linewidth}
			\centering
			\includegraphics[width=\linewidth]{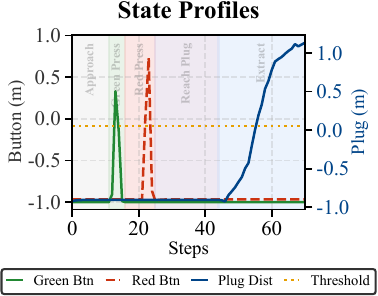} \\
			\vspace{1mm} \small (\arabic{tempcounter}) Task IV \stepcounter{tempcounter}
		\end{minipage}\hfill
		\begin{minipage}{0.2\linewidth}
			\centering
			\includegraphics[width=\linewidth]{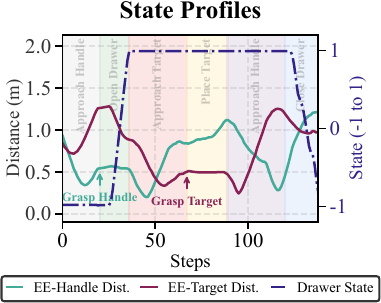} \\
			\vspace{1mm} \small (\arabic{tempcounter}) Task V
		\end{minipage}
	}
	
	\caption{Execution results for the five evaluated tasks. (a) 3D trajectories of the end-effector and manipulated objects. Solid blue lines represent the end-effector path. Dashed lines show the object movement. Circles and squares mark the start positions. Stars and dark squares denote the final states. (b) State transitions and physical profiles for each stage. These include joint angles for Task I and spatial coordinates for Tasks II and III. Task IV shows button activation signals and plug displacement. Task V presents the distance to the target and drawer states.}
	\label{fig:all_tasks}
\end{figure*}



\subsection{Qualitative Analysis of Task Execution}
\label{subsec:qualitative_analysis}
The proposed method is evaluated across five distinct intra-cabin tasks (detailed in Fig.~\ref{fig:env_images}), each designed to challenge specific manipulation capabilities.

For Task I, as shown in Fig.~\ref{fig:all_tasks}(a)(1), the end-effector trajectory aligns closely with the handle displacement post-grasp, ensuring stable manipulation throughout the sequence. The angular profiles in Fig.~\ref{fig:all_tasks}(b)(1) further characterize the task through three distinct phases: Approaching, Rotating, and Opening. Following the initial approach, handle rotation increases sharply while the hatch remains stationary. Hatch actuation initiates only after the handle rotation reaches its predefined threshold, successfully completing the transition from unlocking to opening. 

Task II results characterize the spatial dynamics of stowing a floating object. Fig.~\ref{fig:all_tasks}(a)(2) illustrates the end-effector and target object trajectories, while the axis positions over time are plotted in Fig.~\ref{fig:all_tasks}(b)(2). The operation consists of distinct approach and placement phases, where the end-effector trajectory (solid line) aligns smoothly with the target (dashed line) to counteract microgravity disturbances.

Furthermore, precision-demanding alignment is evaluated through Task III. The three-phase position curves in Fig.~\ref{fig:all_tasks}(b)(3) demonstrate that the end-effector coordinates precisely track the lid across all dimensions. In the final "closing" phase, the trajectories converge smoothly and remain stable. This stability, coupled with the high trajectory overlap between the end-effector and the lid shown in Fig.~\ref{fig:all_tasks}(a)(3), confirms that the controller satisfies the rigorous high-precision alignment requirements during task execution.

Fig.~\ref{fig:all_tasks}(a)(4) illustrates the end-effector trajectory navigating through the target switches to the final plug location for Task IV. The dual Y-axis plot in Fig.~\ref{fig:all_tasks}(b)(4) captures the state transitions for this complex sequence. The left axis records activation peaks during the green and red button presses, while the right axis tracks the displacement during plug removal. These corresponding peaks demonstrate the temporal consistency and precision of the policy across the multi-step "open, close, and pull" sequence.

Finally, Task V characterizes the spatial trajectories of the end-effector and the object during a long-horizon drawer manipulation, as depicted in Fig.~\ref{fig:all_tasks}(a)(5). The state profiles in Fig.~\ref{fig:all_tasks}(b)(5) explicitly reveal the underlying logic of this sequential task. Constant phases in the distance curve (solid line) correspond to the "grasp handle" and "grasp target" actions. Simultaneously, on the right axis, the drawer state signal (dash-dotted line) accurately tracks the physical transitions from closed, to open, and back to closed, validating the framework's capability in extended sequential planning.

\subsection{Energy Consumption Model}
\label{subsec:energy_model}
Energy efficiency is evaluated using a theoretical model based on 45nm CMOS technology \cite{rathi2021diet, rueckauer2017conversion}. The model accounts for two primary operation types: multiply-accumulate (MAC) operations for dense, continuous-value computation and AC operations for sparse, binary spiking computation. The energy costs are defined as $E_{\text{MAC}} = 4.6$~pJ and $E_{\text{AC}} = 0.9$~pJ, respectively.

\subsubsection{Energy Model for ANN-based Policy}
The energy consumption of ANN-based diffusion policies is dominated by dense matrix multiplications. For a policy with $L$ layers and $K$ denoising steps, the total inference energy $E_{\text{ANN}}$ is calculated as:
\begin{equation}
	E_{\text{ANN}} = K \sum_{l=1}^{L} \text{MACs}^{(l)} \cdot E_{\text{MAC}},
\end{equation}
where $\text{MACs}^{(l)}$ denotes the number of MAC operations in layer $l$.

\subsubsection{Energy Model for SNN-based Policy}
To accurately model our spiking residual diffusion network, the architecture is partitioned into static interface layers $\mathcal{L}_{sta}$ and $M$ full-spiking residual blocks. The total energy $E_{\text{SNN}}$ is the sum of static energy $E_{sta}$ and dynamic energy $E_{dyn}$.

Static interface layers process continuous representations and perform MAC operations once per denoising step $k$. The static energy is computed as:
\begin{equation}
	E_{sta} = K \sum_{l \in \mathcal{L}_{sta}} \text{MACs}^{(l)} \cdot E_{\text{MAC}}.
\end{equation}

The dynamic energy $E_{dyn}$ is consumed by the synaptic operations (SOPs) triggered by event-driven spikes. Since the residual block is fully spiking, both the main cascaded modules and the residual path rely exclusively on sparse AC operations. It is formulated as:
\begin{equation}
	E_{dyn} = \sum_{k=1}^{K} \sum_{\tau=1}^{T} \sum_{m=1}^{M} \text{SOPs}_{k, \tau}^{[m]} \cdot E_{\text{AC}},
\end{equation}
where $\text{SOPs}_{k, \tau}^{[m]}$ denotes the total number of AC operations required by the $m$-th block (including both the main path and the residual projection $\mathcal{H}(\cdot)$) at diffusion step $k$ and SNN timestep $\tau$. Note that when the input and output dimensions of the residual block are identical, $\mathcal{H}(\cdot)$ acts as an identity mapping, which contributes no additional AC operations.

\subsection{Performance and Efficiency Evaluation}
\label{subsec:performance_eval}

Table~\ref{tab:performance_results} presents a quantitative comparison between the proposed L-SDPPO method and several baseline algorithms. The experimental results demonstrate that our method almost outperforms all baselines across all tested tasks, achieving a mean SR of $0.97$ and an average AR of $478.25$. This represents a significant margin over the best-performing baseline, DPPO, and substantially exceeds traditional methods such as GBC and PPO. 

Beyond success rates and rewards, energy efficiency is a critical metric for intra-vehicular robotic manipulation. As shown in Table~\ref{tab:performance_results}, the continuous deep neural network baseline, DPPO, improves the average SR to 0.79 but demands a substantial EC of 628.75. This highlights the heavy computational burden of traditional approaches. Our L-SDPPO framework resolves this trade-off effectively. It restricts the average EC to only 229.18, representing roughly 36.45\% of the DPPO cost. Ultimately, SNNs provide a clear efficiency benefit by utilizing temporal sparsity for low-power yet precise control.

\begin{table*}[t]
	\centering
	\caption{Quantitative Comparison of Different Methods Across Multiple Tasks}
	\label{tab:performance_results}
	\scriptsize 
	\renewcommand{\arraystretch}{1.3} 
	\setlength{\tabcolsep}{0pt} 
	
	\begin{tabular*}{\textwidth}{@{\extracolsep{\fill}} l ccc ccc ccc ccc ccc ccc @{}}
		\toprule
		\multirow{2}{*}{Method} & \multicolumn{3}{c}{Task I} & \multicolumn{3}{c}{Task II} & \multicolumn{3}{c}{Task III} & \multicolumn{3}{c}{Task IV} & \multicolumn{3}{c}{Task V} & \multicolumn{3}{c}{Average} \\
		\cmidrule(lr){2-4} \cmidrule(lr){5-7} \cmidrule(lr){8-10} \cmidrule(lr){11-13} \cmidrule(lr){14-16} \cmidrule(lr){17-19}
		& SR & EC & AR & SR & EC & AR & SR & EC & AR & SR & EC & AR & SR & EC & AR & SR & EC & AR \\
		\midrule
		GBC \cite{chernova2007confidence} & 0.25 & --- & 19.00 & 0.32 & --- & 65.18 & 0.00 & --- & 0.00 & 0.00 & --- & 0.00 & 0.03 & --- & 2.58 & 0.12 & --- & 17.35 \\
		DP \cite{chi2025diffusion} & 0.53 & 418.99 & 213.10 & 0.88 & 419.04 & 293.62 & 0.29 & 419.37 & 42.57 & 0.67 & 419.51 & 257.80 & 0.42 & 418.94 & 27.70 & 0.56 & 419.17 & 166.96 \\
		PPO \cite{schulman2017proximal} & 0.88 & --- & 395.88 & 0.00 & --- & 0.00 & 0.00 & --- & 0.00 & 0.00 & --- & 0.00 & \textbf{1.00} & --- & \textbf{583.75} & 0.38 & --- & 195.93 \\
		GPPO \cite{nair2018overcoming} & 0.28 & --- & 97.82 & 0.38 & --- & 70.12 & 0.00 & --- & 0.00 & 0.10 & --- & 20.23 & 0.73 & --- & 99.53 & 0.30 & --- & 57.54 \\
		DPPO \cite{ren2024diffusion}& 1.00 & 628.49 & 604.25 & 0.98 & 628.56 & 394.13 & 0.13 & 629.05 & 10.94 & \textbf{1.00} & 629.26 & \textbf{577.43} & 0.84 & 628.41 & 166.25 & 0.79 & 628.75 & 350.60 \\
		\midrule
		\textbf{L-SDPPO (Ours)} & \textbf{1.00} & \textbf{232.22} & \textbf{658.41} & \textbf{0.99} & \textbf{239.91} & \textbf{414.02} & \textbf{0.91} & \textbf{227.45} & \textbf{502.36} & 0.98 & \textbf{228.87} & 564.09 & 0.98 & \textbf{217.45} & 252.36 & \textbf{0.97} & \textbf{229.18} & \textbf{478.25} \\
		\bottomrule
	\end{tabular*}
\end{table*}

\begin{figure*}[htp]
	\centering
	\begin{minipage}{0.19\textwidth}
		\centering
		\subfloat[]{
			\includegraphics[width=\linewidth]{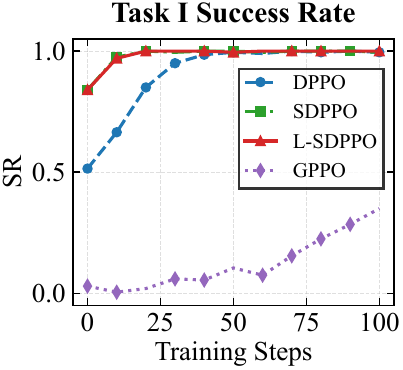}
		}
	\end{minipage}
	\hfill
	\begin{minipage}{0.19\textwidth}
		\centering
		\subfloat[]{
			\includegraphics[width=\linewidth]{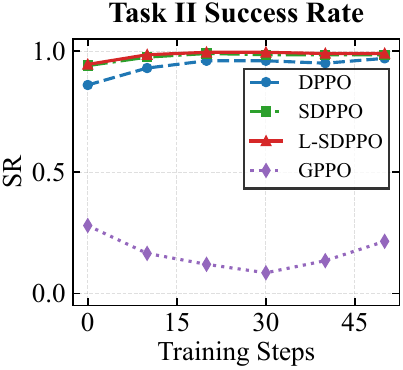}
		}
	\end{minipage}
	\hfill
	\begin{minipage}{0.19\textwidth}
		\centering
		\subfloat[]{
			\includegraphics[width=\linewidth]{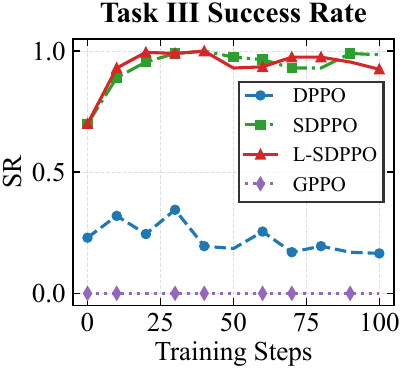}
		}
	\end{minipage}
	\hfill
	\begin{minipage}{0.19\textwidth}
		\centering
		\subfloat[]{
			\includegraphics[width=\linewidth]{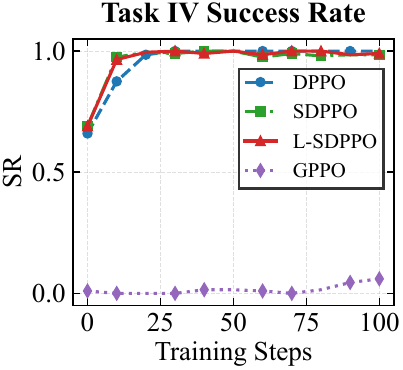}
		}
	\end{minipage}
	\hfill
	\begin{minipage}{0.19\textwidth}
		\centering
		\subfloat[]{
			\includegraphics[width=\linewidth]{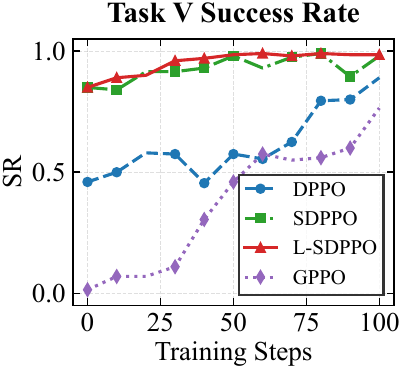}
		}
	\end{minipage}
	
	\caption{Success rate comparison across five different tasks.}
	\label{fig:success_rates}
\end{figure*}

\begin{figure*}[htp]
	\centering
	\begin{minipage}{0.19\textwidth}
		\centering
		\subfloat[]{
			\includegraphics[width=\linewidth]{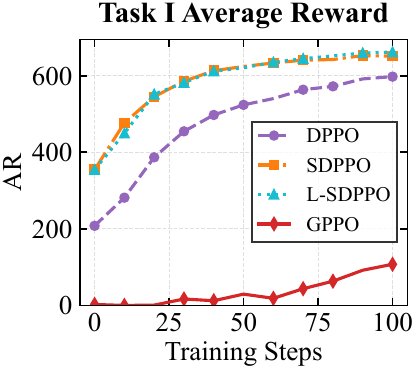}
		}
	\end{minipage}
	\hfill
	\begin{minipage}{0.19\textwidth}
		\centering
		\subfloat[]{
			\includegraphics[width=\linewidth]{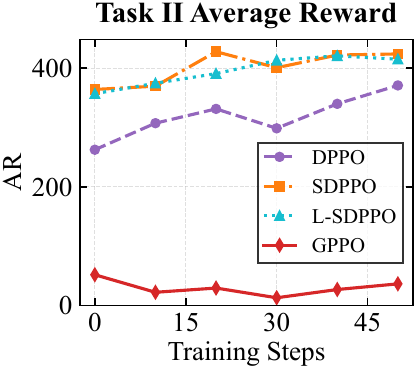}
		}
	\end{minipage}
	\hfill
	\begin{minipage}{0.19\textwidth}
		\centering
		\subfloat[]{
			\includegraphics[width=\linewidth]{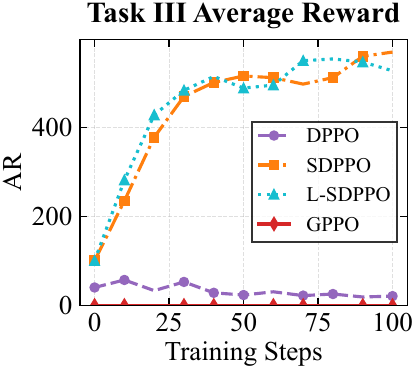}
		}
	\end{minipage}
	\hfill
	\begin{minipage}{0.19\textwidth}
		\centering
		\subfloat[]{
			\includegraphics[width=\linewidth]{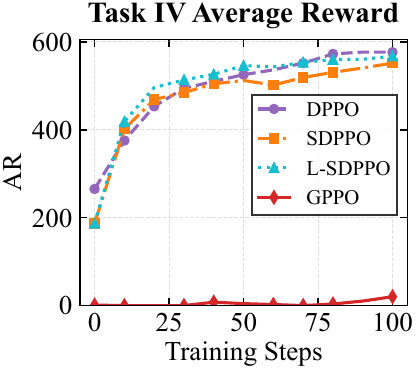}
		}
	\end{minipage}
	\hfill
	\begin{minipage}{0.19\textwidth}
		\centering
		\subfloat[]{
			\includegraphics[width=\linewidth]{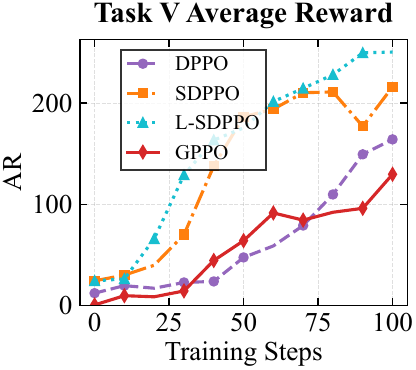}
		}
	\end{minipage}
	
	\caption{Average reward comparison across five different tasks.}
	\label{fig:average_rewards}
\end{figure*}

\begin{table*}[t] 
	\centering
	\caption{Ablation Study of the Proposed Method Across Multiple Tasks}
	\label{tab:ablation_study}
	\scriptsize 
	\renewcommand{\arraystretch}{1.3} 
	\setlength{\tabcolsep}{1.2pt} 
	
	\begin{tabular*}{\textwidth}{@{\extracolsep{\fill}} lcccccccccccc @{}}
		\toprule
		\multirow{2}{*}{Method} & \multicolumn{2}{c}{Task I} & \multicolumn{2}{c}{Task II} & \multicolumn{2}{c}{Task III} & \multicolumn{2}{c}{Task IV} & \multicolumn{2}{c}{Task V} & \multicolumn{2}{c}{Avg.} \\
		\cmidrule(lr){2-3} \cmidrule(lr){4-5} \cmidrule(lr){6-7} \cmidrule(lr){8-9} \cmidrule(lr){10-11} \cmidrule(lr){12-13}
		& SR & AR & SR & AR & SR & AR & SR & AR & SR & AR & SR & AR \\
		\midrule
		SDP & 0.84 & 349.36 & 0.92 & 354.48 & 0.67 & 105.86 & 0.70 & 185.01 & 0.78 & 46.00 & 0.78 & 208.14 \\
		SDPPO & 1.00 & 652.18 & 0.92 & 400.13 & 0.83 & 457.68 & 0.98 & 544.35 & 0.98 & 209.90 & 0.94 & 452.85 \\
		\midrule
		\textbf{L-SDPPO} & \textbf{1.00} & \textbf{658.41} & \textbf{0.99} & \textbf{414.02} & \textbf{0.91} & \textbf{502.36} & \textbf{0.98} & \textbf{564.09} & \textbf{0.98} & \textbf{252.36} & \textbf{0.97} & \textbf{478.25} \\
		\bottomrule
	\end{tabular*}
\end{table*}


Figures \ref{fig:success_rates} and \ref{fig:average_rewards} plot the SR and AR during training. The SNN-based methods, SDPPO and L-SDPPO, consistently converge faster and achieve higher final metrics than the ANN baselines (GPPO and DPPO). This performance gap is particularly evident in the precision container capping task (Task III). Here, the ANN baselines struggle to find viable policies and remain at low success rates. Conversely, the SNN methods quickly stabilize at success rates above 0.9. The reward curves further show that while DPPO can learn, its growth rate and final rewards fall significantly short of the SNN approaches.

\begin{figure}[t]
	\centering
	
	\subfloat[GBC]{
		\includegraphics[width=0.23\linewidth]{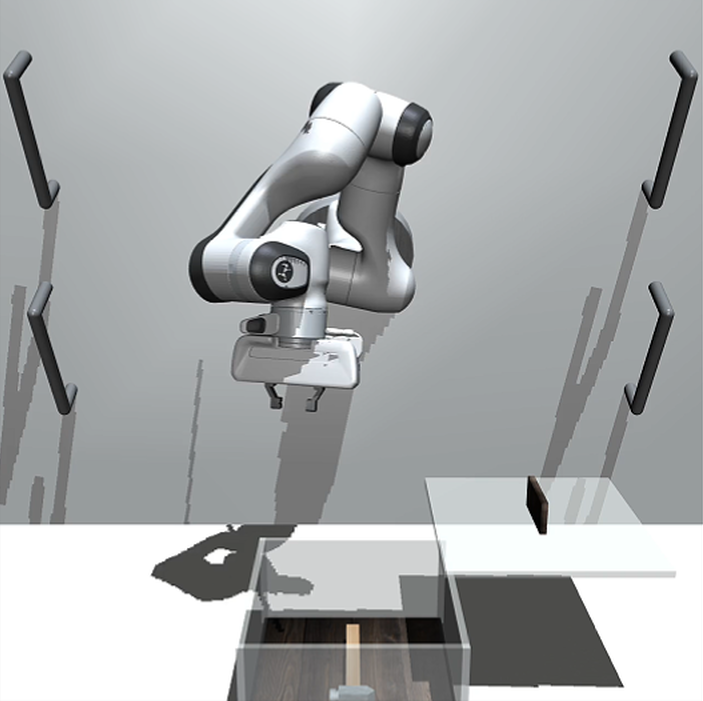} \hfill
		\includegraphics[width=0.23\linewidth]{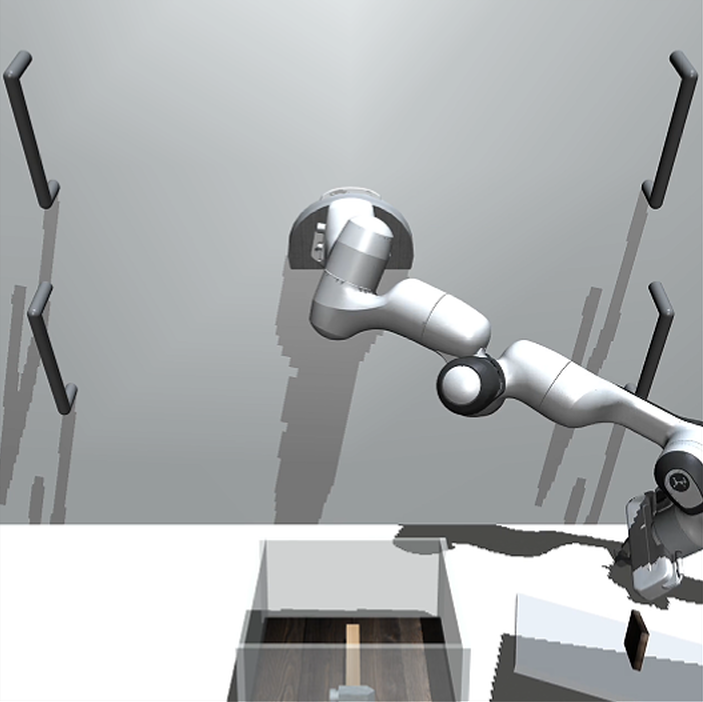} \hfill
		\includegraphics[width=0.23\linewidth]{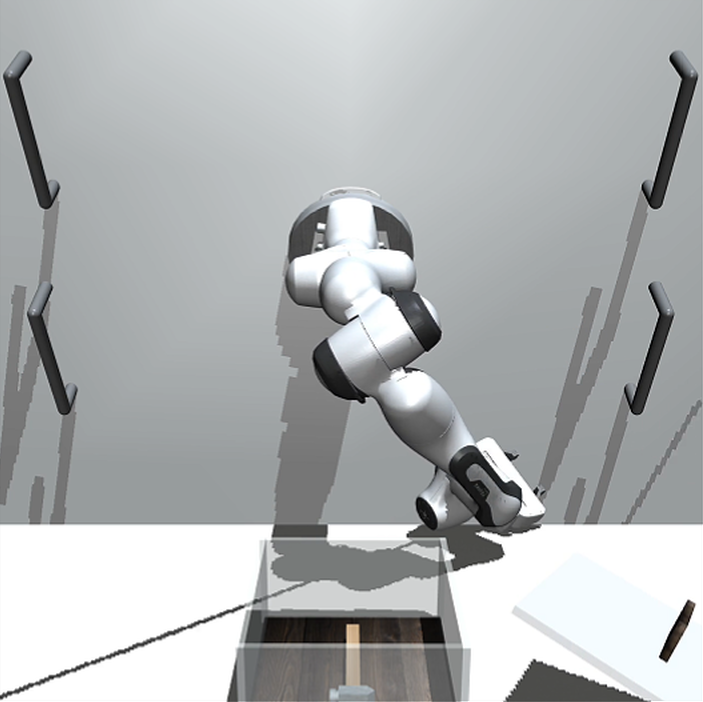} \hfill
		\includegraphics[width=0.23\linewidth]{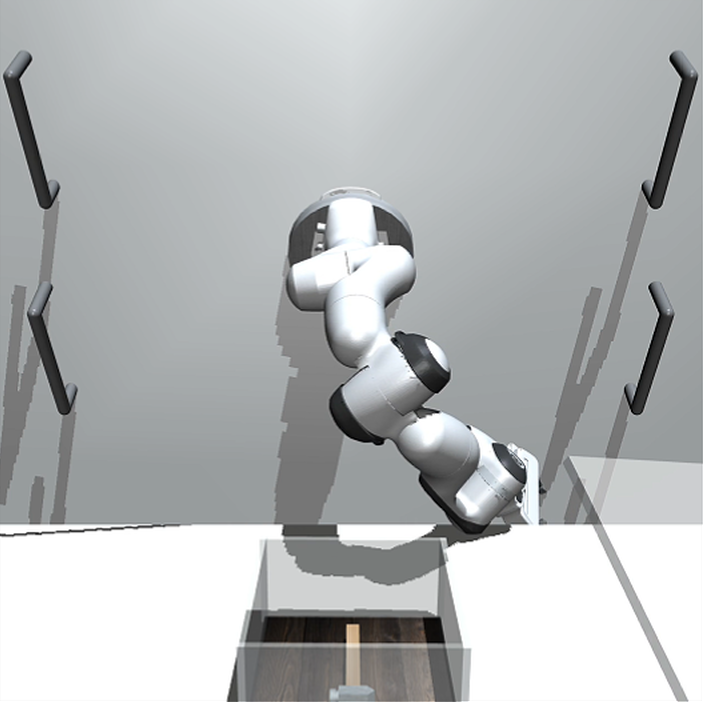}
	}

	\subfloat[DP]{
		\includegraphics[width=0.23\linewidth]{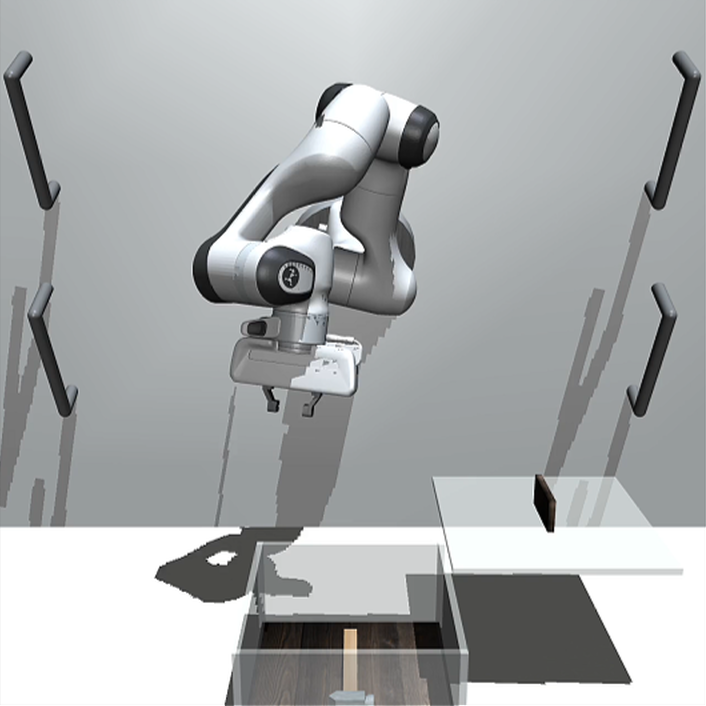} \hfill
		\includegraphics[width=0.23\linewidth]{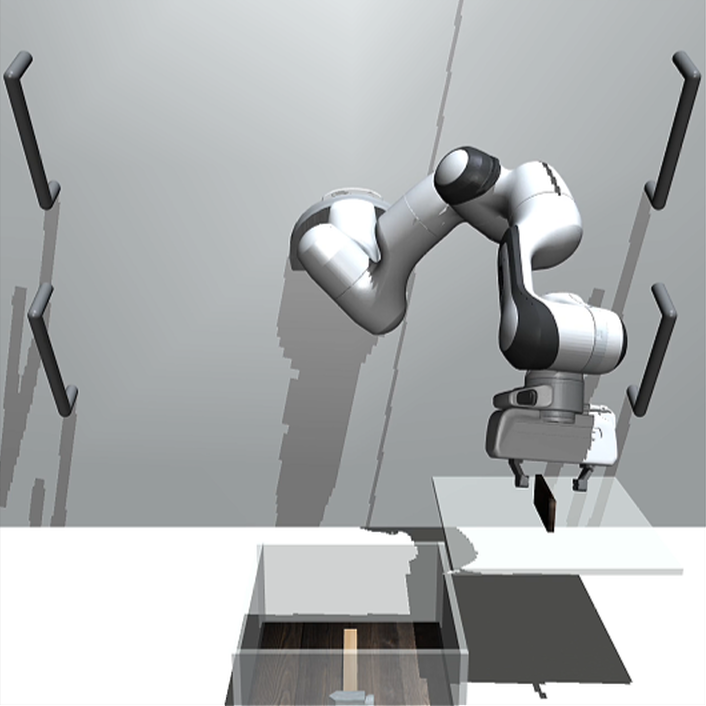} \hfill
		\includegraphics[width=0.23\linewidth]{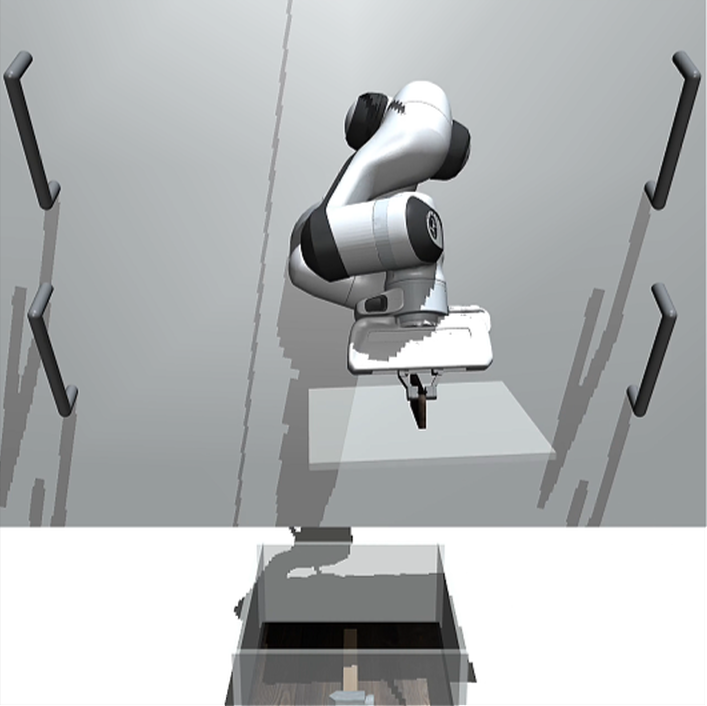} \hfill
		\includegraphics[width=0.23\linewidth]{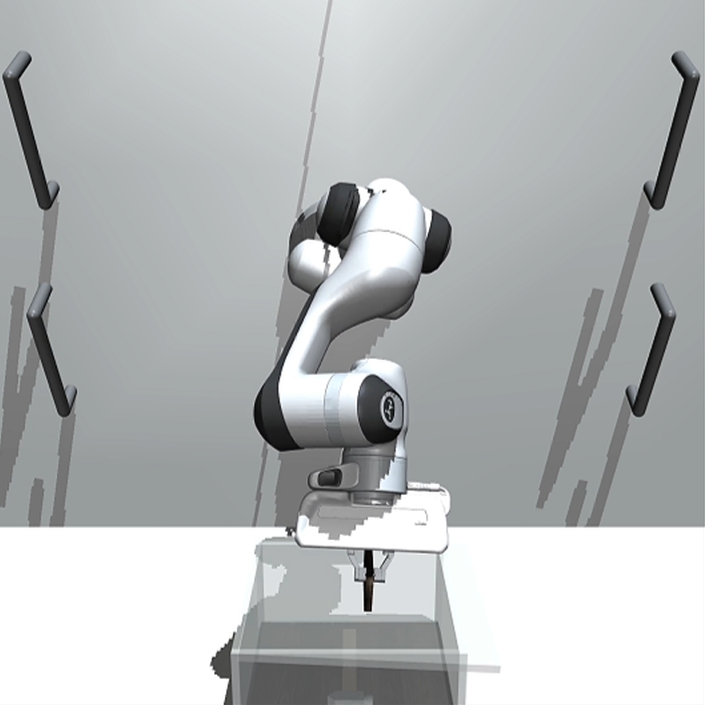}
	}

	\subfloat[PPO]{
		\includegraphics[width=0.23\linewidth]{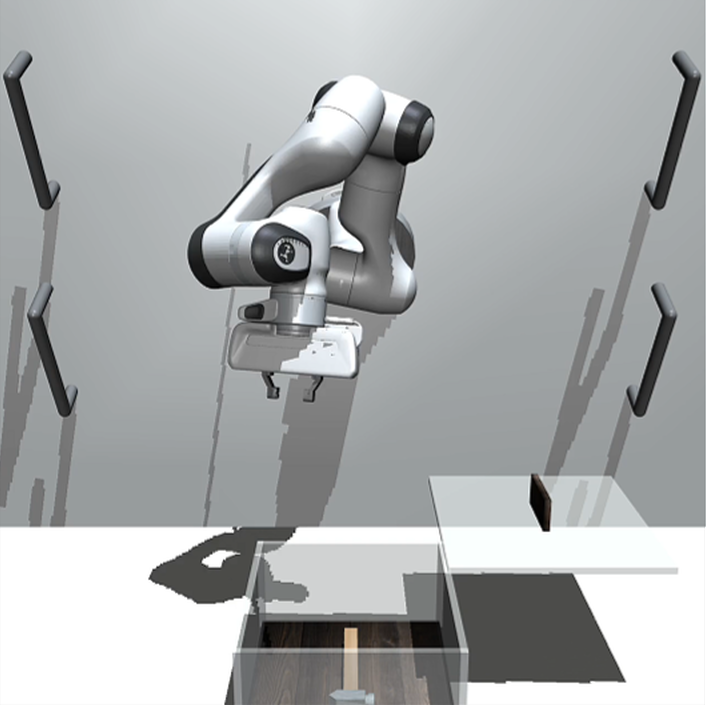} \hfill
		\includegraphics[width=0.23\linewidth]{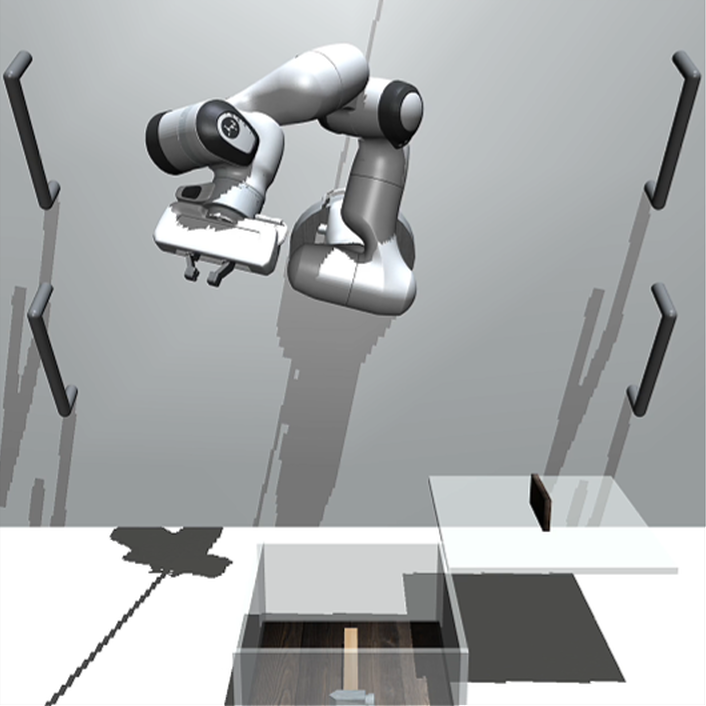} \hfill
		\includegraphics[width=0.23\linewidth]{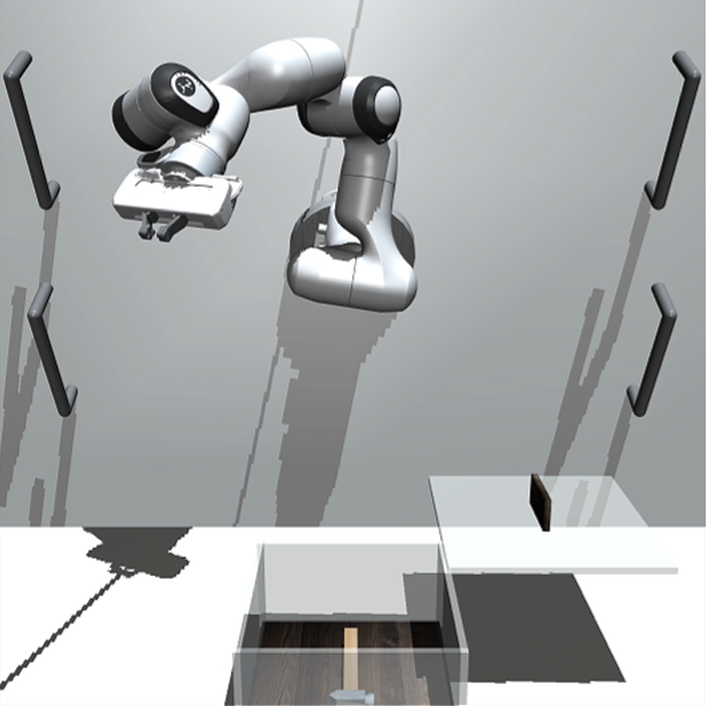} \hfill
		\includegraphics[width=0.23\linewidth]{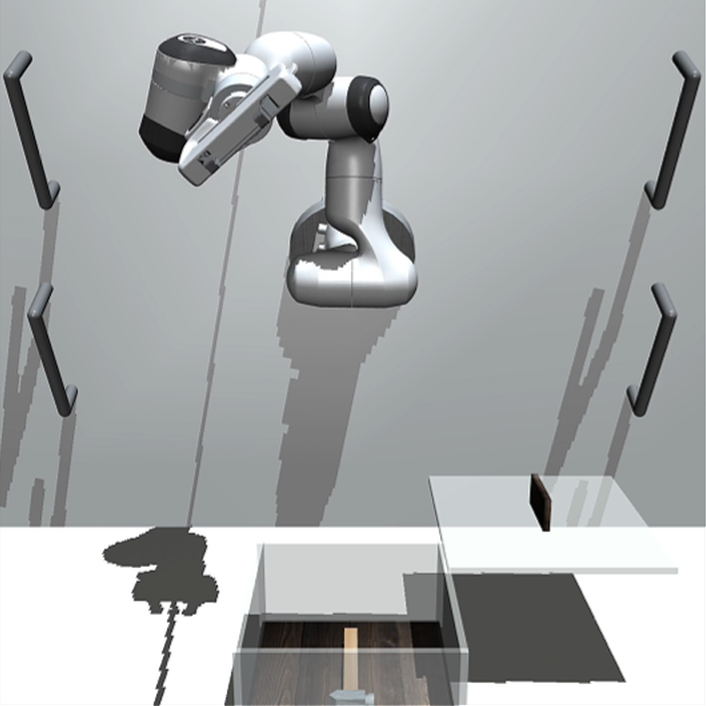}
	}

	\subfloat[GPPO]{
		\includegraphics[width=0.23\linewidth]{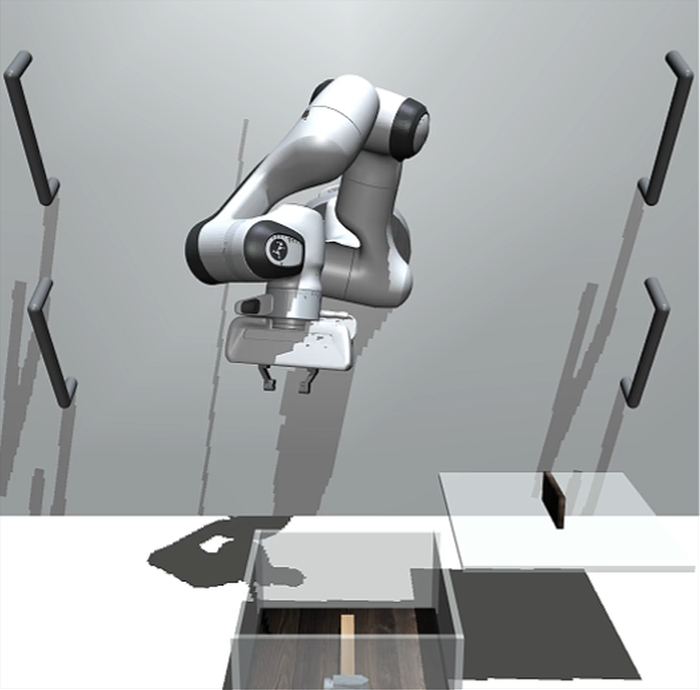} \hfill
		\includegraphics[width=0.23\linewidth]{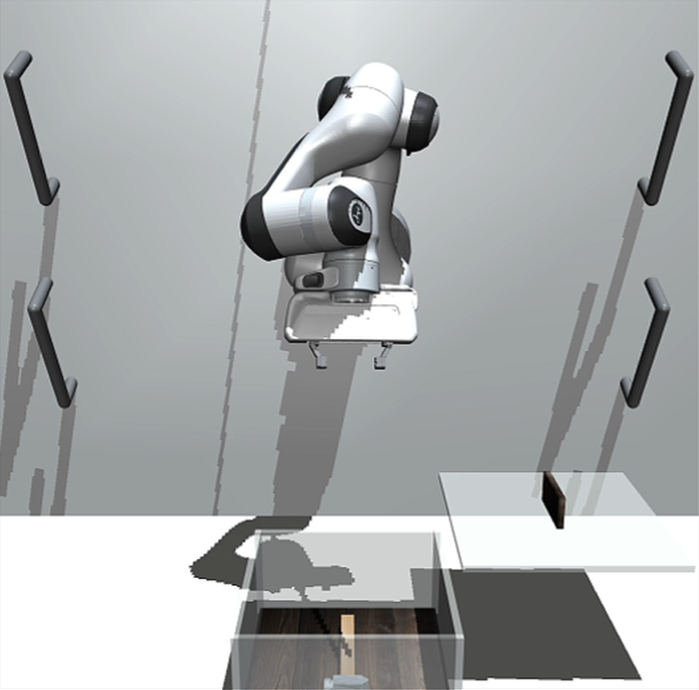} \hfill
		\includegraphics[width=0.23\linewidth]{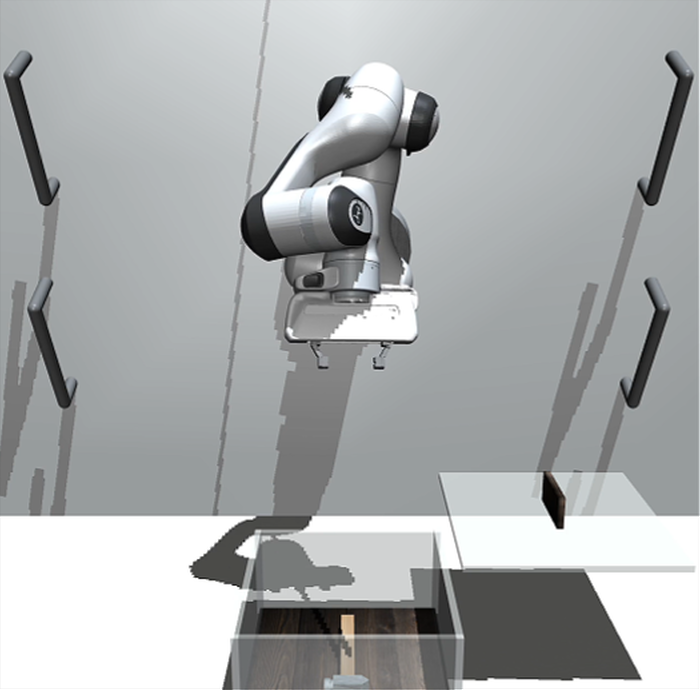} \hfill
		\includegraphics[width=0.23\linewidth]{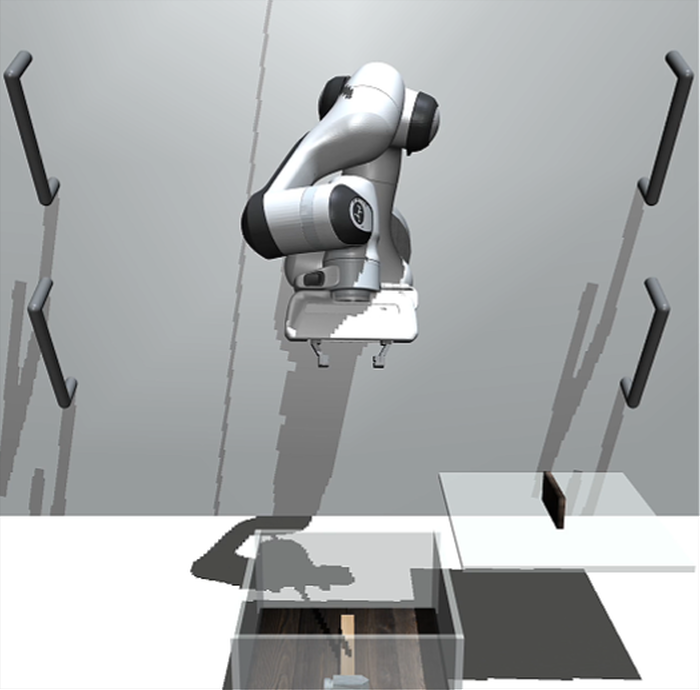}
	}
	
	\subfloat[DPPO]{
		\includegraphics[width=0.23\linewidth]{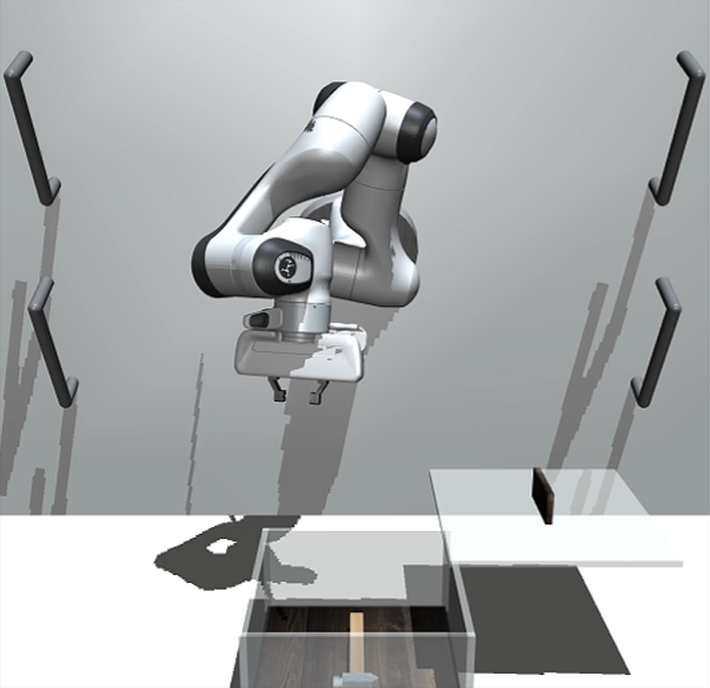} \hfill
		\includegraphics[width=0.23\linewidth]{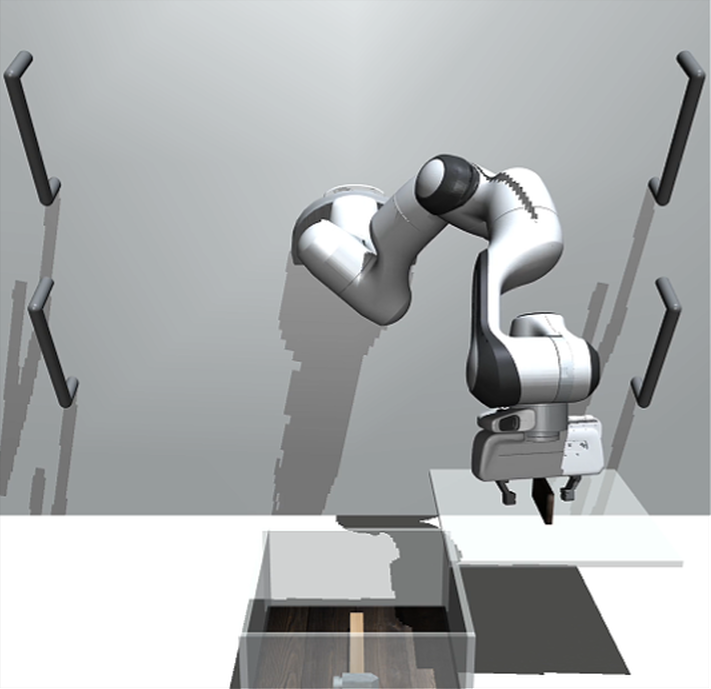} \hfill
		\includegraphics[width=0.23\linewidth]{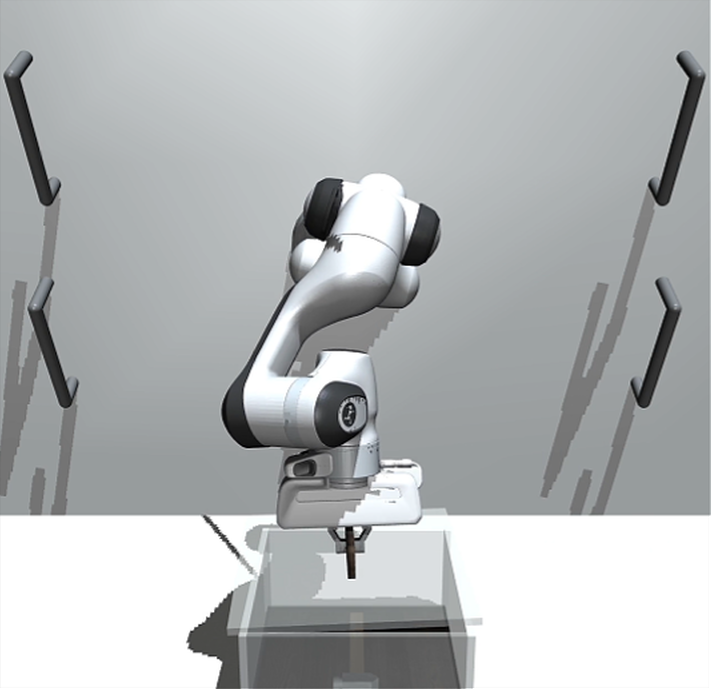} \hfill
		\includegraphics[width=0.23\linewidth]{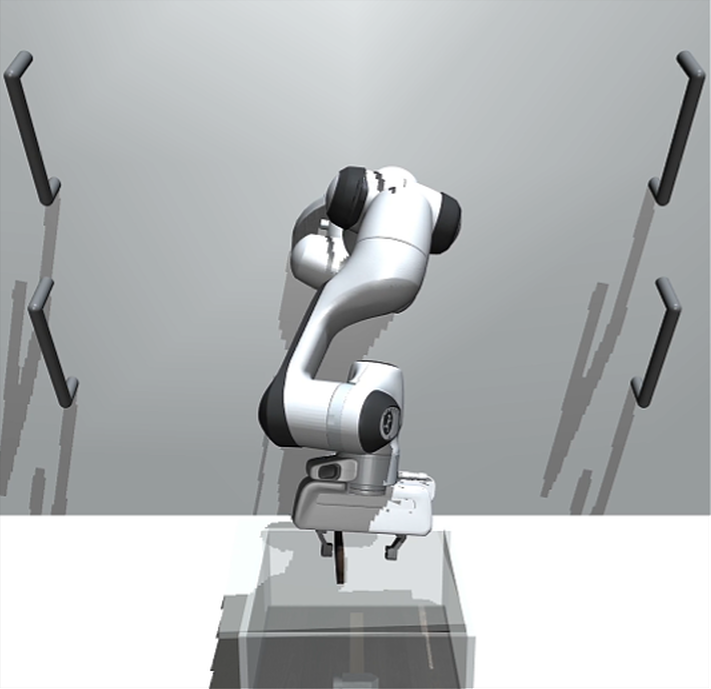}
	}
	
	\subfloat[L-SDPPO]{
	\includegraphics[width=0.23\linewidth]{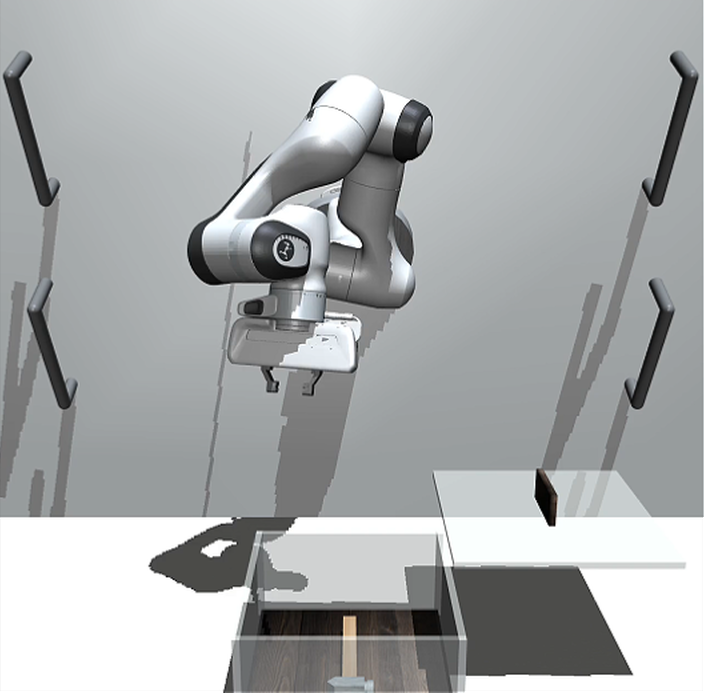} \hfill
	\includegraphics[width=0.23\linewidth]{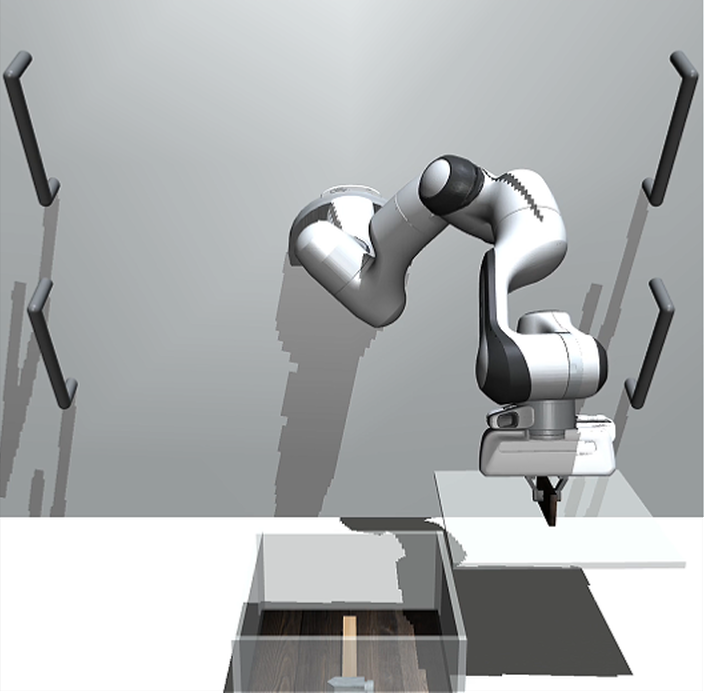} \hfill
	\includegraphics[width=0.23\linewidth]{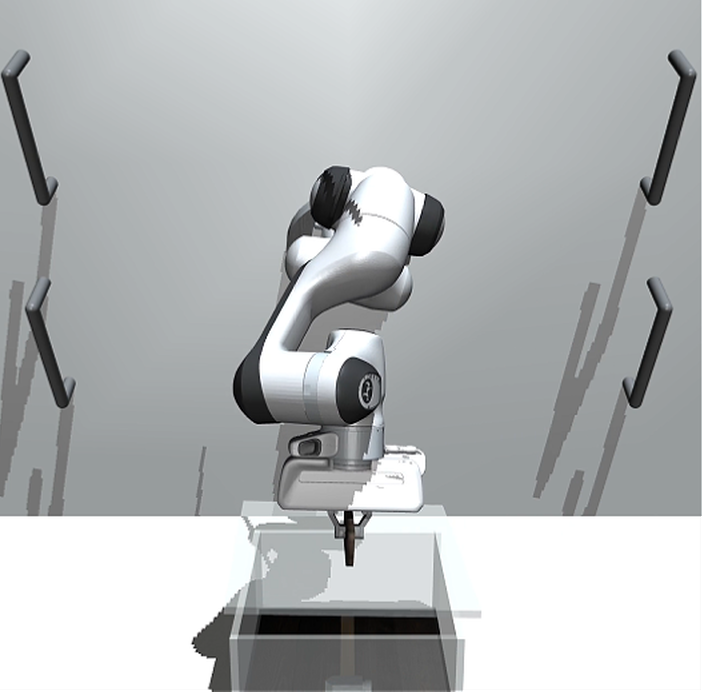} \hfill
	\includegraphics[width=0.23\linewidth]{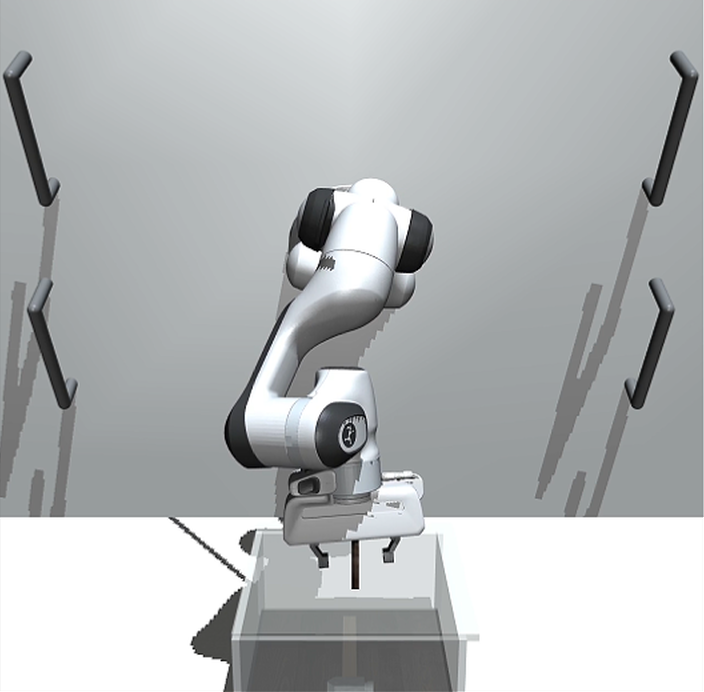}
}
	
	\caption{Qualitative comparison of the execution sequences for Task III across different algorithms.}
	\label{fig_comparison}
\end{figure}

To visually corroborate these quantitative findings, Fig.~\ref{fig_comparison} illustrates the execution sequences for Task III across all evaluated algorithms. As observed in the progressive frames, traditional baselines such as GBC and PPO often exhibit erratic trajectories or severe overshooting, completely failing to align with the target. While the continuous DPPO baseline manages to approach the container, it visibly oscillates during the final millimeter-level alignment phase. In contrast, our proposed L-SDPPO framework maintains a stable, smooth, and highly precise spatial trajectory throughout the entire capping process.These results definitively indicate that the spatio-temporal dynamics of SNNs provide better representation precision for fine-grained control tasks.

\subsection{Ablation Study}
\label{subsec:ablation_study}

To systematically evaluate the contribution of each core component in our proposed framework, we conduct an ablation study by isolating the reinforcement learning fine-tuning phase and the SDLI mechanism. We define three distinct policy variants for evaluation:
\begin{itemize}
	\item SDP: The base SNN-driven diffusion model trained exclusively via behavior cloning on expert demonstrations. This variant lacks both RL fine-tuning and the latency injection mechanism.
	\item SDPPO: The SDP model further fine-tuned using the PPO algorithm to maximize environment rewards. However, the SDLI module is ablated, meaning the SNN operates with fixed temporal dynamics.
\end{itemize}

The ablation results in Table~\ref{tab:ablation_study} validate the contribution of each proposed module. Relying solely on IL pre-training limits the SDP baseline to an average SR of 0.78 and an AR of 208.14. The addition of PPO fine-tuning in the SDPPO model provides a substantial upgrade. It elevates the average SR to 0.94 and brings the AR to 452.85. Building upon this, our full L-SDPPO method achieves the highest overall metrics with an average SR of 0.97 and an AR of 478.25. The value of the SDLI module is most prominent in the high-precision Box Closing scenario of Task III. Compared to the SDPPO model, adding SDLI pushes the Task III SR from 0.83 to 0.91 and increases the AR from 457.68 to 502.36. This confirms that SDLI is crucial for precise and stable control.

\section{Conclusion}

This paper addresses the strict power constraints and complex manipulation requirements of intra-cabin robots in microgravity environments. We propose L-SDPPO, an integrated neuromorphic generative control framework that synergizes spiking diffusion policies with online reinforcement learning. To fundamentally resolve the challenges of dynamic spatio-temporal feature perception, we introduce a SDLI mechanism to dynamically regulate the temporal dynamics of the spiking network. Extensive evaluations across five space cabin simulation tasks demonstrate that our method achieves a superior average success rate of 0.97, outperforming conventional ANN baselines. Furthermore, by leveraging the event-driven sparsity of SNNs, L-SDPPO restricts inference energy consumption to approximately 36.45\% of that required by traditional continuous diffusion models. This optimal balance of high-precision control and drastic computational reduction makes the L-SDPPO framework highly practical for power-limited spacecraft.

Future work will evaluate the algorithm on physical neuromorphic hardware to verify real-world energy consumption. We also plan to extend the framework to bimanual coordination and soft-object manipulation for complex on-orbit servicing missions.


\bibliographystyle{IEEEtran}
\bibliography{References}  

\begin{thebibliography}{10}
\providecommand{\url}[1]{#1}
\csname url@samestyle\endcsname
\providecommand{\newblock}{\relax}
\providecommand{\bibinfo}[2]{#2}
\providecommand{\BIBentrySTDinterwordspacing}{\spaceskip=0pt\relax}
\providecommand{\BIBentryALTinterwordstretchfactor}{4}
\providecommand{\BIBentryALTinterwordspacing}{\spaceskip=\fontdimen2\font plus
\BIBentryALTinterwordstretchfactor\fontdimen3\font minus
  \fontdimen4\font\relax}
\providecommand{\BIBforeignlanguage}[2]{{%
\expandafter\ifx\csname l@#1\endcsname\relax
\typeout{** WARNING: IEEEtran.bst: No hyphenation pattern has been}%
\typeout{** loaded for the language `#1'. Using the pattern for}%
\typeout{** the default language instead.}%
\else
\language=\csname l@#1\endcsname
\fi
#2}}
\providecommand{\BIBdecl}{\relax}
\BIBdecl

\bibitem{russell2006applying}
J.~F. Russell, D.~M. Klaus, and T.~J. Mosher, ``Applying analysis of
  international space station crew-time utilization to mission design,''
  \emph{Journal of spacecraft and rockets}, vol.~43, no.~1, pp. 130--136, 2006.

\bibitem{li2022review}
C.~Li, J.~Yang, and S.~Chang, ``Review on key technologies of space intelligent
  grasping robot,'' \emph{Journal of the Brazilian Society of Mechanical
  Sciences and Engineering}, vol.~44, no.~2, p.~64, 2022.

\bibitem{perez2018velocity}
P.~R. P{\'e}rez, M.~De~Stefano, and R.~Lampariello, ``Velocity matching
  compliant control for a space robot during capture of a free-floating
  target,'' in \emph{Proceedings of the 2018 IEEE Aerospace Conference}, Big
  Sky, MT, USA, 2018, pp. 1--9.

\bibitem{ma2023advances}
B.~Ma, Z.~Jiang, Y.~Liu, and Z.~Xie, ``Advances in space robots for on-orbit
  servicing: A comprehensive review,'' \emph{Advanced Intelligent Systems},
  vol.~5, no.~8, p. 2200397, 2023.

\bibitem{capra2024learning}
L.~Capra, M.~D'Ambrosio, and M.~Lavagna, ``Learning-based trajectory
  optimization of a space manipulator posttarget-grasping,'' in
  \emph{Proceedings of the 75th International Astronautical Congress}, Milan,
  Italy, 2024, pp. 361--370.

\bibitem{hou2020data}
Z.~Hou, J.~Fei, Y.~Deng, and J.~Xu, ``Data-efficient hierarchical reinforcement
  learning for robotic assembly control applications,'' \emph{IEEE Transactions
  on Industrial Electronics}, vol.~68, no.~11, pp. 11\,565--11\,575, 2020.

\bibitem{li2025research_aca}
M.~Li, Y.~Huang, H.~Zhang \emph{et~al.}, ``Research on grasping and
  transferring floating objects by space robots using combined
  imitation-reinforcement learning,'' in \emph{Proceedings of the 23rd IFAC
  Symposium on Automatic Control in Aerospace (ACA)}, ser. IFAC-PapersOnLine,
  vol.~59, no.~20.\hskip 1em plus 0.5em minus 0.4em\relax Harbin, China:
  Elsevier, 2025, pp. 1545--1550.

\bibitem{wang2024novel}
W.~Wang, C.~Zeng, Z.~Lu, and C.~Yang, ``A novel robust imitation learning
  framework for dual-arm object-moving tasks,'' \emph{IEEE Transactions on
  Industrial Electronics}, vol.~71, no.~12, pp. 16\,068--16\,076, 2024.

\bibitem{hua2021learning}
J.~Hua, L.~Zeng, G.~Li, and Z.~Ju, ``Learning for a robot: Deep reinforcement
  learning, imitation learning, transfer learning,'' \emph{Sensors}, vol.~21,
  no.~4, p. 1278, 2021.

\bibitem{chi2025diffusion}
C.~Chi, Z.~Xu, S.~Feng, E.~Cousineau, Y.~Du, B.~Burchfiel, R.~Tedrake, and
  S.~Song, ``Diffusion policy: Visuomotor policy learning via action
  diffusion,'' \emph{The International Journal of Robotics Research}, vol.~44,
  no. 10-11, pp. 1684--1704, 2025.

\bibitem{briden2025diffusion}
J.~Briden, B.~J. Johnson, R.~Linares, and A.~Cauligi, ``Diffusion policies for
  generative modeling of spacecraft trajectories,'' in \emph{AIAA SCITECH 2025
  Forum}, Orlando, FL, USA, 2025, p. 2775.

\bibitem{tavanaei2019deep}
A.~Tavanaei, M.~Ghodrati, S.~R. Kheradpisheh, T.~Masquelier, and A.~Maida,
  ``Deep learning in spiking neural networks,'' \emph{Neural networks}, vol.
  111, pp. 47--63, 2019.

\bibitem{eshraghian2023training}
J.~K. Eshraghian, M.~Ward, E.~O. Neftci, X.~Wang, G.~Lenz, G.~Dwivedi,
  M.~Bennamoun, D.~S. Jeong, and W.~D. Lu, ``Training spiking neural networks
  using lessons from deep learning,'' \emph{Proceedings of the IEEE}, vol. 111,
  no.~9, pp. 1016--1054, 2023.

\bibitem{zhang2026multimodal}
L.~Zhang, G.~Sun, and H.~Deng, ``Multimodal spiking neural network for space
  robotic manipulation,'' \emph{Acta Astronautica}, 2026.

\bibitem{wang2014emg}
N.~Wang, C.~Yang, M.~R. Lyu, and Z.~Li, ``An emg enhanced impedance and force
  control framework for telerobot operation in space,'' in \emph{Proceedings of
  the 2014 IEEE Aerospace Conference}, Big Sky, MT, USA, 2014, pp. 1--10.

\bibitem{platt2011multiple}
R.~Platt, M.~Abdallah, and C.~Wampler, ``Multiple-priority impedance control,''
  in \emph{Proceedings of the IEEE International Conference on Robotics and
  Automation}, Shanghai, China, 2011, pp. 6033--6038.

\bibitem{palma2023application}
P.~Palma, T.~Rybus, and K.~Seweryn, ``Application of impedance control of the
  free floating space manipulator for removal of space debris,'' \emph{Pomiary
  Automatyka Robotyka}, vol.~27, no.~3, pp. 95--106, 2023.

\bibitem{liu2019research}
D.~Liu, H.~Liu, Y.~Liu, and Z.~Li, ``Research on impedance control of flexible
  joint space manipulator on-orbit servicing,'' in \emph{Proceedings of the
  2019 IEEE International Conference on Robotics and Biomimetics}, Dali, China,
  2019, pp. 77--82.

\bibitem{peng2024reinforcement}
Z.~Peng and C.~Wang, ``Reinforcement learning-based pose coordination planning
  capture strategy for space non-cooperative targets,'' \emph{Aerospace},
  vol.~11, no.~9, p. 706, 2024.

\bibitem{hu2025deep}
Y.~Hu, D.~Zhou, W.~Yao, X.~Shao, and G.~Sun, ``Deep reinforcement
  learning-based trajectory planning with continuous pose representation for
  6-dof free-floating space robot,'' \emph{Aerospace Science and Technology},
  vol. 166, p. 110540, 2025.

\bibitem{li2021constrained}
Y.~Li, X.~Hao, Y.~She, S.~Li, and M.~Yu, ``Constrained motion planning of
  free-float dual-arm space manipulator via deep reinforcement learning,''
  \emph{Aerospace Science and Technology}, vol. 109, p. 106446, 2021.

\bibitem{al2024path}
A.~Al~Ali, J.-F. Shi, and Z.~H. Zhu, ``Path planning of 6-dof free-floating
  space robotic manipulators using reinforcement learning,'' \emph{Acta
  Astronautica}, vol. 224, pp. 367--378, 2024.

\bibitem{blaise2023space}
J.~Blaise and M.~C. Bazzocchi, ``Space manipulator collision avoidance using a
  deep reinforcement learning control,'' \emph{Aerospace}, vol.~10, no.~9, p.
  778, 2023.

\bibitem{nair2018overcoming}
A.~Nair, B.~McGrew, M.~Andrychowicz, W.~Zaremba, and P.~Abbeel, ``Overcoming
  exploration in reinforcement learning with demonstrations,'' in
  \emph{Proceedings of the IEEE International Conference on Robotics and
  Automation}, Brisbane, Australia, 2018, pp. 6292--6299.

\bibitem{vecerik2017leveraging}
M.~Vecerik, T.~Hester, J.~Scholz, F.~Wang, O.~Pietquin, B.~Piot, N.~Heess,
  T.~Roth{\"o}rl, T.~Lampe, and M.~Riedmiller, ``Leveraging demonstrations for
  deep reinforcement learning on robotics problems with sparse rewards,''
  \emph{arXiv preprint arXiv:1707.08817}, 2017.

\bibitem{shao2026imitation}
S.~Shao, D.~Zhou, G.~Sun, L.~Zhang, and M.~Jiang, ``Imitation learning-based
  spacecraft rendezvous and docking method with expert demonstration,''
  \emph{arXiv preprint arXiv:2601.12952}, 2026.

\bibitem{shyam2020imitation}
R.~Shyam, Z.~Hao, U.~Montanaro, and G.~Neumann, ``Imitation learning for
  autonomous trajectory learning of robot arms in space,'' \emph{arXiv preprint
  arXiv:2008.04007}, 2020.

\bibitem{ashith2021autonomous}
R.~Ashith~Shyam, Z.~Hao, U.~Montanaro, S.~Dixit, A.~Rathinam, Y.~Gao,
  G.~Neumann, and S.~Fallah, ``Autonomous robots for space: Trajectory learning
  and adaptation using imitation,'' \emph{Frontiers in Robotics and AI},
  vol.~8, p. 638849, 2021.

\bibitem{joukov2017gaussian}
V.~Joukov and D.~Kulic, ``Gaussian process based model predictive controller
  for imitation learning,'' in \emph{2017 IEEE-RAS 17th International
  Conference on Humanoid Robotics (Humanoids)}, Birmingham, UK, 2017, pp.
  850--855.

\bibitem{ning2025integrated}
Y.~Ning, T.~Li, Y.~Zhang, Z.~Li, W.~Du, and Y.~Zhang, ``An integrated framework
  of grasp detection and imitation learning for space robotics applications,''
  \emph{Chinese Journal of Mechanical Engineering}, vol.~38, no.~1, p. 139,
  2025.

\bibitem{janner2022planning}
M.~Janner, Y.~Du, J.~B. Tenenbaum, and S.~Levine, ``Planning with diffusion for
  flexible behavior synthesis,'' \emph{arXiv preprint arXiv:2205.09991}, 2022.

\bibitem{ze20243d}
Y.~Ze, G.~Zhang, K.~Zhang, C.~Hu, M.~Wang, and H.~Xu, ``3d diffusion policy:
  Generalizable visuomotor policy learning via simple 3d representations,''
  \emph{arXiv preprint arXiv:2403.03954}, 2024.

\bibitem{team2024octo}
O.~M. Team, D.~Ghosh, H.~Walke, K.~Pertsch, K.~Black, O.~Mees, S.~Dasari,
  J.~Hejna, T.~Kreiman, C.~Xu \emph{et~al.}, ``Octo: An open-source generalist
  robot policy,'' \emph{arXiv preprint arXiv:2405.12213}, 2024.

\bibitem{ma2024hierarchical}
X.~Ma, S.~Patidar, I.~Haughton, and S.~James, ``Hierarchical diffusion policy
  for kinematics-aware multi-task robotic manipulation,'' in \emph{Proceedings
  of the IEEE/CVF Conference on Computer Vision and Pattern Recognition},
  Seattle, WA, USA, 2024, pp. 18\,081--18\,090.

\bibitem{yu2025d3p}
S.-A. Yu, F.~Gao, Y.~Wu, C.~Yu, and Y.~Wang, ``D3p: Dynamic denoising diffusion
  policy via reinforcement learning,'' \emph{arXiv preprint arXiv:2508.06804},
  2025.

\bibitem{ren2024diffusion}
A.~Z. Ren, J.~Lidard, L.~L. Ankile, A.~Simeonov, P.~Agrawal, A.~Majumdar,
  B.~Burchfiel, H.~Dai, and M.~Simchowitz, ``Diffusion policy policy
  optimization,'' \emph{arXiv preprint arXiv:2409.00588}, 2024.

\bibitem{chernova2007confidence}
S.~Chernova and M.~Veloso, ``Confidence-based policy learning from
  demonstration using gaussian mixture models,'' in \emph{Proceedings of the
  6th International Joint Conference on Autonomous Agents and Multiagent
  Systems}, Honolulu, Hawaii, USA, 2007, pp. 1--8.

\bibitem{schulman2017proximal}
J.~Schulman, F.~Wolski, P.~Dhariwal, A.~Radford, and O.~Klimov, ``Proximal
  policy optimization algorithms,'' \emph{arXiv preprint arXiv:1707.06347},
  2017.

\bibitem{rathi2021diet}
N.~Rathi and K.~Roy, ``Diet-snn: A low-latency spiking neural network with
  direct input encoding and leakage and threshold optimization,'' \emph{IEEE
  Transactions on Neural Networks and Learning Systems}, vol.~34, no.~6, pp.
  3174--3182, 2021.

\bibitem{rueckauer2017conversion}
B.~Rueckauer, I.-A. Lungu, Y.~Hu, M.~Pfeiffer, and S.-C. Liu, ``Conversion of
  continuous-valued deep networks to efficient event-driven networks for image
  classification,'' \emph{Frontiers in neuroscience}, vol.~11, p. 682, 2017.

\end{thebibliography}
\end{document}